\documentclass[acmsmall,screen,nonacm]{acmart}
\copyrightyear{2026}
\acmYear{2026}

\usepackage{multirow}
\usepackage{numprint}

\usepackage{algorithm}
\usepackage{algpseudocode}

\usepackage[table]{xcolor}

\npdecimalsign{.}

\begin{document}

\title[Privacy-Preserving Distributed Optimization Using MPC and Evolutionary Algorithms]{Privacy-Preserving Distributed Optimization Under Time Constraints Using Secure Multi-Party Computation and Evolutionary Algorithms}

\author{Sebastian Gruber}
\email{sebastian.gruber@jku.at}
\orcid{0000-0002-5714-1870}
\author{Tobias Harzfeld}
\email{tobias.harzfeld@jku.at}
\orcid{0009-0003-0275-6207}
\author{Christoph G. Schuetz}
\email{christoph.schuetz@jku.at}
\orcid{0000-0002-0955-8647}

\affiliation{%
  \institution{Johannes Kepler University Linz}
  \city{Linz}
  \country{Austria}
}

\author{Florian Wohner}
\email{florian.wohner@ait.ac.at}
\orcid{0000-0002-8641-7522}
\author{Thomas Lor\"unser}
\email{thomas.loruenser@ait.ac.at}
\orcid{0000-0002-1829-4882}

\affiliation{%
  \institution{AIT Austrian Institute of Technology}
  \city{Vienna}
  \country{Austria}
}

\renewcommand{\shortauthors}{Gruber et al.}

\begin{abstract}
  In distributed optimization, multiple parties collaborate to find an optimal solution to a problem. Privacy-preserving distributed optimization uses techniques, such as secure multi-party computation (MPC), to protect the private inputs of each party. In time-critical settings, the runtime overhead introduced by privacy-preserving computations may prevent the optimization from finishing within the deadline. This paper presents an approach for privacy-preserving distributed optimization in time-critical settings that combines evolutionary algorithms for solution search and MPC for the evaluation of solutions. The approach reduces the impact of privacy-preserving computations on runtime and allows to return solution within the deadline. Obfuscation of evaluation results provides additional protection for private inputs from an honest-but-curious platform provider, but introduces a potential trade-off between protection and solution quality. This trade-off is investigated in experiments using a genetic algorithm for both the single-objective assignment problem and the traveling salesperson problem, as well as NSGA-II for the multi-objective assignment problem.
\end{abstract}

\begin{CCSXML}
<ccs2012>
   <concept>
       <concept_id>10002950.10003624.10003625.10003630</concept_id>
       <concept_desc>Mathematics of computing~Combinatorial optimization</concept_desc>
       <concept_significance>300</concept_significance>
       </concept>
   <concept>
       <concept_id>10002978.10003022.10003028</concept_id>
       <concept_desc>Security and privacy~Domain-specific security and privacy architectures</concept_desc>
       <concept_significance>500</concept_significance>
       </concept>
   <concept>
       <concept_id>10010147.10010257.10010293.10011809.10011812</concept_id>
       <concept_desc>Computing methodologies~Genetic algorithms</concept_desc>
       <concept_significance>300</concept_significance>
       </concept>
 </ccs2012>
\end{CCSXML}

\ccsdesc[500]{Security and privacy~Domain-specific security and privacy architectures}
\ccsdesc[300]{Computing methodologies~Genetic algorithms}
\ccsdesc[300]{Mathematics of computing~Combinatorial optimization}

\keywords{evolutionary computation, privacy-preserving optimization, time-critical optimization, multi-objective optimization}

\maketitle

\section{Introduction}

In distributed optimization, multiple parties collaborate to find an optimal solution to a problem.
In this paper, we consider distributed optimization problems in which no single party possesses all the information required to find a solution (see~\cite{sakuma2007,lu2018}).
For example, in air traffic flow management, when congestion occurs, flights must be delayed, with delay costs varying across flights and some flights being less critical to their airlines than others~\cite{pilon2021}.
Optimizing the flight list in such cases requires knowledge of the delay cost functions for each affected flight, which is known only to the respective airline.

Privacy-preserving distributed optimization protects the private information of the parties that contribute to solving an optimization problem, e.g., using (secure) multi-party computation (MPC) via secret-sharing~\cite{cramer2015} or homomorphic encryption~\cite{yi2014}.
For example, the delay cost functions for flights affected by congestion are confidential, and using MPC or homomorphic encryption allows optimization of the flight list without requiring airlines to reveal this confidential information to other airlines or to a potentially not fully trusted third party.

For time-critical tasks, in which the ``correctness of a computation depends not only on the logical correctness but also on the time at which the results are produced''~\cite[p.~6]{shin1994}, the ``software routine implementing the functionality of a task should complete its execution before the task deadline''~\cite[p.~8]{mitra2018}.
Hence, under strict time constraints, a heuristic optimization algorithm that delivers a sufficiently good solution within the deadline may be preferable to an exact algorithm that produces an optimal solution too late.
For example, in air traffic flow management, flight lists must be optimized within a limited timeframe during congestion to allow timely verification and implementation of the updated schedules.

Privacy-preserving computation inevitably introduces runtime overhead, which may render privacy-preserving optimization impractical in time-critical settings.
For example, in air traffic flow management, privacy-preserving implementations of exact optimization algorithms~\cite{loruenser2022} may fail to find an optimal flight list before the deadline.

In this paper, we propose an approach for privacy-preserving distributed optimization in time-critical settings, in which an \emph{Optimizer} component uses an evolutionary algorithm to search for solutions, while a \emph{Privacy Engine} component employs MPC only to evaluate the solutions, returning obfuscated evaluation results that indicate relative fitness of the solutions.
Compared to a fully privacy-preserving implementation of the optimization algorithm, this selective use of MPC reduces computational overhead, resulting in lower runtime.
Moreover, an evolutionary algorithm can be terminated at any time and still provide a valid solution.

We assume an honest-but-curious platform provider that receives encrypted confidential inputs that can be used for solving combinatorial optimization problems.
While following the protocol, the platform provider may attempt to infer information from the received data.
Privacy-preserving computation protects these confidential inputs from such a platform provider, but this protection comes with a potential trade-off in solution quality due to the use of heuristic optimization in connection with obfuscation methods that conceal exact fitness values from the evolutionary algorithm.
In this paper, we experimentally investigate this trade-off for the single-objective assignment problem (AP), the multi-objective assignment problem (MOAP), and the traveling salesperson problem (TSP).

The remainder of this paper is organized as follows.
In Section~\ref{sec:related}, we review related work.
In Section~\ref{sec:approach}, we present the proposed approach.
In Section~\ref{sec:setup}, we describe the experimental setup for evaluation.
In Section~\ref{sec:results}, we present and interpret the experimental results.
We conclude the paper with Section~\ref{sec:conclusion}.

\section{Related Work}\label{sec:related}

In this section, we discuss related work on privacy-preserving optimization using evolutionary algorithms.
Privacy-preserving optimization often uses cryptographic techniques, such as MPC~\cite{cramer2015} and homomorphic encryption~\cite{yi2014}, which enable computation over private data while protecting the data.
MPC enables multiple parties to jointly compute functions using secret-sharing, while homomorphic encryption allows computing functions on encrypted data.
These techniques enable privacy-preserving computations without a trusted third party, but they incur additional runtime overhead.
Privacy-preserving implementations of exact algorithms may therefore not be able to provide a solution within the time constraints of a given application.

Evolutionary algorithms may not find optimal solutions, but are capable of providing valid solutions at any time.
Combining evolutionary algorithms and cryptographic techniques for privacy-preserving computation enables the search for efficient solutions while satisfying time constraints and protecting private information.
Various approaches for privacy-preserving optimization using evolutionary algorithms have already been proposed in the literature.

Sakuma and Kobayashi ~\shortcite{sakuma2007} propose a protocol for using a local search and a genetic algorithm to solve combinatorial problems such as the TSP involving two parties.
They distinguish between a server, which holds the cost function, and a searcher, which holds the set of cities.
Han and Ng~\shortcite{han2007} propose a protocol for using a genetic algorithm to discover decision rules in a dataset distributed between two parties.
Sakuma and Kobayashi~\shortcite{sakuma2007} reveal the relative order of solutions to the searcher, whereas Han and Ng~\shortcite{han2007} reveal exact fitness values.

Zhan et al.~\shortcite{zhan2021} propose a framework for evolutionary computation involving two parties.
The framework distinguishes between a user, who holds the objective function, and a designer, who performs the evolutionary operations.
The user evaluates the populations produced by the designer, but applies a rank-based cryptographic function to only reveal the ranks to the designer.
Sun et al.~\shortcite{sun2023} extend this framework for privacy-preserving feature selection.

She et al.~\shortcite{she2023} propose a similar approach for bi-party multi-objective optimization involving a server, which performs the evolutionary operations, and two decision makers, who evaluate populations with their own objective functions.
The decision makers reveal the Pareto ranks and crowding distances of the solutions to the server.

Liu et al.~\shortcite{liu2024b} argue that, beyond fitness values, the ranking itself can reveal private information.
They propose a federated data-driven evolutionary algorithm for multi-objective optimization using a Diffie-Hellman-assisted secure aggregation.
This approach protects both the predicted fitness values and the ranking of individuals.

Jiang and Fu~\shortcite{jiang2020} combine a genetic algorithm and homomorphic encryption to allow a user to outsource optimization to a server.
The server performs the optimization and returns an encrypted optimization result to the user.

Zhao et al.~\shortcite{zhao2024} propose evolutionary computation as a service, and implement a genetic algorithm for combinatorial optimization.
The idea is to allow a user to outsource optimization to a server that performs evolutionary operations on encrypted data and returns an encrypted optimization result to the user.
Funke and Kerschbaum~\shortcite{funke2010} propose an evolutionary algorithm for multi-objective optimization with two parties, each party aiming to optimize its own objective function.
This approach only reveals the final optimization result to both parties.
Both approaches secure the entire algorithm.

Our approach assumes distributed optimization problems in time-critical settings.
In contrast to approaches that secure the entire algorithm~\cite{funke2010, zhao2024}, our approach uses methods for privacy-preserving computation only for the evaluation of the solutions to reduce the impact of privacy-preserving computations on runtime.
In contrast to approaches that reveal results of the objective functions~\cite{sakuma2007, han2007, zhan2021, she2023}, our approach obfuscates evaluation results to reduce the information that is revealed to an honest-but-curious platform provider.
In this paper, we present our approach and experimentally evaluate the impact of obfuscation on the performance of evolutionary algorithms.

\section{Approach}\label{sec:approach}

In this section, we present our approach for privacy-preserving distributed optimization under time constraints, combining MPC with evolutionary algorithms.
We extend the approach proposed by Schuetz et al.~\shortcite{schuetz2022a}, originally developed for solving single-objective flight prioritization using MPC in combination with genetic algorithms, to a wider class of potentially multi-objective optimization problems.
The original approach has also been applied successfully to single-objective flight prioritization using MPC with local search algorithms~\cite{schuetz2023} as well as to multi-objective flight prioritization using MPC with multi-objective genetic algorithms~\cite{gruber2024a}.
However, these studies have been limited to the domain-specific setting of air traffic management.
We argue that the approach is broadly applicable to distributed optimization problems arising in real-world settings with privacy concerns and time constraints.

\subsection{Problem Formulation}

We consider the following class of distributed, possibly multi-objective optimization problems, in which the information that defines the objectives is distributed between multiple parties.
We adapt the definitions of Ehrgott~\shortcite{ehrgott2005} for multi-objective optimization and the definitions of Yang et al.~\shortcite{yang2019} for distributed optimization.
Let $\mathcal{X}$ be the set of feasible solutions for a multi-objective optimization problem with a vector
\[
f(x) = \big(f_1(x), \dots, f_n(x)\big), \quad x \in \mathcal{X},
\]
consisting of $n$ objective functions.
The goal is to find optimal solutions with respect to the vector of objective functions, i.e., solutions $x^\star \in \mathcal{X}$ for which there is no $x \in \mathcal{X}$, $x \neq x^\star$, such that
\[
\theta(f(x)) \preceq \theta(f(x^\star))
\]
where $\theta$ is a mapping of the space of the vector of objective functions to a space for which an order relation $\preceq$ is defined.
Let $p$ denote the number of parties. 
For each objective function $f_i(x)$, $ i = 1, \dots, n$, each party possesses a private component $f_{ij}(x)$, $j = 1, \dots, p$.
The global objective functions are then defined as
\[
f_i(x) = \sum_{j=1}^p f_{ij}(x), \quad i = 1, \dots, n.
\]
No single party has access to all $f_{ij}(x)$ or to the full vector of objective functions $f(x)$, so the optimization problem is distributed in the sense that no single party possesses sufficient information to independently evaluate $f(x)$.

We implement and investigate the proposed approach for combinatorial optimization problems, where a solution can be represented by a binary matrix.
In particular, we instantiate the generic definition for the single-objective and multi-objective \emph{linear sum assignment problem} as well as for the \emph{traveling salesperson problem}.

Let $A = (a_{ij})$ be an $m \times n$ matrix representing a candidate solution, where $a_{ij} \in \{0, 1\}$ for all $i \in \{1,\dots,m\}$ and $j \in \{1,\dots,n\}$.
Here, each feasible solution $x \in \mathcal{X}$ is represented by a binary matrix $A$.
For linear sum assignment problems, the matrix encodes assignments of items from a set $U$ to items of a set $V$, with $|U| = m$ and $|V| = n$; both balanced ($m = n$) and imbalanced ($m < n$) cases are admissible.
In contrast, for the TSP, the matrix encodes transitions between items of the same set, i.e., $U = V$, and thus $m = n$.
Feasible solutions are defined by problem-specific constraints on $A$.

Let $C = (c_{ij})$ denote a global \emph{weight matrix}, where $c_{ij}$ represents the utility or cost associated with setting $a_{ij} = 1$.
This induces an objective function of the form
\[
f_1(A) = \sum_{i=1}^m \sum_{j=1}^n a_{ij} c_{ij},
\]
which is to be maximized or minimized, depending on whether $c_{ij}$ encodes utility or cost.
For multi-objective problems, each objective function $f_i$ is defined analogously via its own global weight matrix $C_i$.

In a distributed setting, there are $p$ parties, where each party $k \in \{1, \dots, p\}$ has access to only a subset of the rows of $C$, corresponding to a subset of items $U_k \subseteq U$, such that $U_1 \cup \dots \cup U_p = U$ and the sets are pairwise disjoint.

\subsection{Architecture}

\begin{figure}
    \centering
    \includegraphics[width=0.9\linewidth]{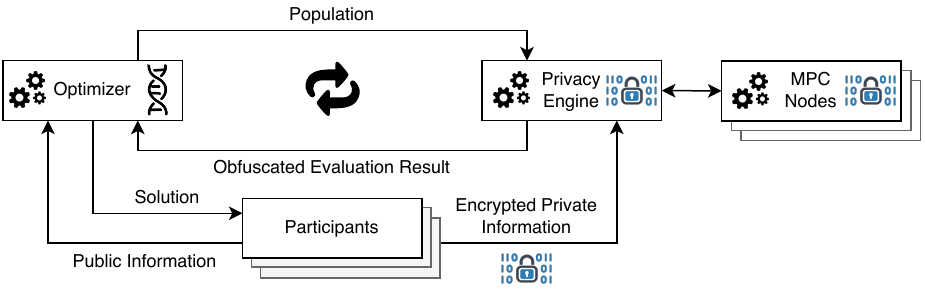}
    \caption{Architecture for privacy-preserving distributed optimization under time constraints}
    \label{fig:framework}
\end{figure}

The proposed architecture for solving distributed optimization problems under time constraints in the presence of an honest-but-curious platform provider separates the search for solutions from their evaluation, with each task performed by a dedicated component.
Figure~\ref{fig:framework} illustrates the interactions between the components of the proposed architecture, with the \emph{Optimizer} and the \emph{Privacy Engine} as central elements responsible for solving the optimization problem.
The Optimizer employs an evolutionary algorithm to generate candidate solutions, while the Privacy Engine evaluates these solutions using MPC.
The separation of solution search and evaluation allows for MPC to be applied only to the evaluation step, reducing runtime overhead compared to optimization algorithms implemented entirely with MPC.

The architecture is designed to limit the information that the platform provider can infer about the optimization preferences of the participants.
The participants provide the Optimizer with public, non-confidential information about the optimization problem (e.g., the items to be assigned), while the Privacy Engine receives the private information defining the optimization preferences (i.e., the objective function) in encrypted form.
Each participant submits the portion of the weight matrix corresponding to the items under the participant's control.
The Privacy Engine has no direct access to the participants' preferences, which are instead distributed across multiple MPC nodes via secret sharing such that no single party has access to the complete information.
The MPC nodes are hosted independently of the platform provider, e.g., by (a subset of) the participants.
Before submitting their portion of the weight matrix, each participant splits their portion into shares (one for each MPC node) and encrypts each share using the public key of the respective MPC node that receives that share.
To further reduce information leakage, the Privacy Engine returns only obfuscated evaluation results, which the Optimizer uses to estimate solution quality and guide the search process.

The architecture seamlessly extends to multi-objective optimization problems, where objective functions may be either public or confidential.
Participants provide the Optimizer with public information for both the problem definition and any public objectives, while confidential objectives are handled by separate instances of the Privacy Engine.
For each confidential objective, a dedicated instance of the Privacy Engine is initialized, receiving encrypted shares of the corresponding private data from a subset of the participants.
The Optimizer evaluates candidate solutions with respect to public objectives directly, whereas confidential objectives are evaluated securely by their respective Privacy Engine instances.

\subsection{Obfuscation Methods}

Even when only evaluation results are shared, the provider may infer relationships between components of the private data.
Consider, for example, a linear sum assignment with a $5 \times 5$ weight matrix $C=(c_{ij})$ to be maximized, with two candidate solutions $x$ and $y$ encoded as vectors (0, 1, 2, 3, 4) and (0, 2, 1, 3, 4), respectively.
The utility of each $x_i$ and $y_i$ then corresponds to $c_{i,x_{i}}$ and $c_{i,y_{i}}$, respectively. 
The fitness values of the solutions are the sums of these utilities.
Suppose now that the solutions $x$ and $y$ have fitness values of 100 and 90, respectively.
From these evaluations, the platform provider can derive the relation 
\[
  c_{1,1} + c_{2,2} = c_{1,2} + c_{2,1} + 10,
\]
since the two solutions differ only at positions 1 and 2, and the fitness value of $x$ exceeds that of $y$ by 10.
Multiple iterations with different populations of candidate solutions allow the platform provider to learn more such pairwise relations between solutions. 
For instance, with 500 iterations and a population size of 300, the provider could construct up to
\[
  \binom{500\times300}{2} = \numprint{11249925000}
\]
equations, assuming no duplicate solutions are generated.
To mitigate this information leakage, Schuetz et al.~\shortcite{schuetz2022a} propose multiple obfuscation methods for evaluation results, which we formalize and further investigate in our optimization setting.

The Privacy Engine employs obfuscation methods to limit the amount of information revealed to an honest-but-curious platform provider during the optimization process.
Hence, the Privacy Engine returns obfuscated fitness values to the Optimizer instead of the actual fitness values.
These obfuscated values are derived from the actual fitness values.
The Optimizer then uses the obfuscated fitness values to guide the selection of solutions.
Formally, to obfuscate the fitness of a solution with respect to an objective $i$, the selection in each generation is guided by a population-dependent mapping $r_i^P: P \rightarrow \mathcal{Y}_i$ instead of directly using the objective function $f_i: \mathcal{X} \rightarrow \mathbb{R}$, which assigns a fitness value to each solution in the set $\mathcal{X}$ of feasible solutions.  
Here, $P \subseteq \mathcal{X}$ is the current population, and the codomain $\mathcal{Y}_i$ depends on the specific obfuscation method used for objective $i$.
We propose the following obfuscation methods, each characterized by a different computation of the mapping $r_i^P$.

\begin{itemize}
    \item \textbf{Order.} The Privacy Engine sorts the solutions of a population by their actual fitness for an objective $i$ and returns the rank of each solution as the obfuscated fitness value.
    Formally, the mapping is a rank function $r_i^P: P \rightarrow\mathbb{N}$ that is monotonic with respect to $f_i$ and assigns equal ranks to equal fitness values. Formally, for any two solutions $x,y \in P$:
    \[
    f_i(x) > f_i(y) \Rightarrow r_i^P(x) < r_i^P(y),
    \]
    with ties broken arbitrarily to obtain a dense ranking.

    \item \textbf{Order Quantiles.}  The Privacy Engine sorts the solutions of a population by their actual fitness for an objective $i$, assigns the sorted solutions to $k$ equal-sized quantiles, and returns the quantile assignment of each solution as the obfuscated fitness value. 
    Formally, let $(x_1, \ldots, x_n)$ denote the population $P$ ordered by fitness such that  $f_i(x_1) \ge f_i(x_2) \ge \ldots \ge f_i(x_n)$.  
    The ordered solutions are then divided into $k$ consecutive groups (quantiles) $Q_1, \ldots, Q_k$, with each group containing approximately $\frac{n}{k}$ solutions, with ties broken arbitrarily.
    The population-dependent mapping $r_i^P : P \to \{1, \ldots, k\}$ is defined by setting $r_i^P(x) = j$ if $x \in Q_j$, returning the quantile index as the obfuscated fitness value.

    \item \textbf{Fitness Buckets.} 
    The Privacy Engine divides the range between the best and worst actual fitness values of a population into $k$ equal-sized buckets, assigns each solution to the bucket covering its actual fitness, and returns the bucket assignment for each solution as the obfuscated fitness value.
    Formally, let $(x_1, \ldots, x_n)$ denote the population $P$, with fitness values $f_i(x_1), \ldots, f_i(x_n)$. 
    Let $f_i^{\max} = \max\limits_{x \in P} f_i(x)$ and $f_i^{\min} = \min\limits_{x \in P} f_i(x)$ denote the best and worst fitness in the population, respectively. 
    The fitness range $[f_i^{\min}, f_i^{\max}]$ is then divided into $k$ equal-sized, non-overlapping intervals (buckets) $B_1, \ldots, B_k$ of length $\frac{f_i^{\max} - f_i^{\min}}{k}$.  
    Each solution $x \in P$ is assigned to the bucket that contains its fitness value $f_i(x)$.
    The population-dependent mapping $r_i^P : P \to \{1, \ldots, k\}$ is defined by setting $r_i^P(x) = j$ if $f_i(x) \in B_j$, returning the bucket index as the obfuscated fitness value.

    \item \textbf{Top Individuals.} The Privacy Engine sorts the solutions of a population by their actual fitness for an objective $i$ and returns the top $k$ solutions.
    Formally, let $(x_1, \ldots, x_n)$ denote the population $P$ ordered by fitness such that  $f_i(x_1) \ge f_i(x_2) \ge \ldots \ge f_i(x_n)$.
    The top $k$ solutions $x_1, \ldots, x_k$ are then selected as the top individuals.  
    The population-dependent mapping $r_i^P : P \to \mathbb{B}$ is defined by
    \[
    r_i^P(x) =
    \begin{cases}
      \text{true} & \text{if } x \text{ is among the top } k \text{ solutions},\\
      \text{false} & \text{otherwise,}
    \end{cases}
    \]
    returning a Boolean value indicating whether a solution is a top individual.
    To bolster obfuscation, the Privacy Engine implementation always returns exactly $k$ individuals, with ties solved arbitrarily.

    \item \textbf{Above Threshold.} The Privacy Engine calculates the range between the best and worst actual fitness values of a population and returns the solutions with fitness at least $k$~\% of this range.
    Formally, let $(x_1, \ldots, x_n)$ denote the population $P$, with fitness values $f_i(x_1), \ldots, f_i(x_n)$.
    Let $f_i^{\max} = \max\limits_{x \in P} f_i(x)$ and $f_i^{\min} = \min\limits_{x \in P} f_i(x)$ denote the best and worst fitness in the population, respectively.
    A fitness threshold $\tau_i$ is then calculated as follows:
    \[
    \tau_i = f_i^{\min} + k \% \times (f_i^{\max} - f_i^{\min}),
    \] 
    where $k \in [0,100]$ is the threshold percentage.
    The population-dependent mapping $r_i^P : P \to \mathbb{B}$ is defined by
    \[
    r_i^P(x) =
    \begin{cases}
      \text{true} & \text{if } f_i(x) \ge \tau_i,\\
      \text{false} & \text{otherwise,}
    \end{cases}
    \]
    returning a Boolean value indicating whether a solution exceeds the fitness threshold.
    To bolster obfuscation, the Privacy Engine implementation always returns at least three individuals, even if not all of them surpass the threshold.
\end{itemize}

The Privacy Engine also returns the maximum actual fitness for each population, allowing the Optimizer to determine whether the solutions improve over time.
However, the identity of the best individual is not revealed to the Optimizer, except when using \emph{order} obfuscation.

Table~\ref{tab:obfuscation_estimation} illustrates the obfuscation methods for a population of eleven solutions: $O$ denotes \emph{order} obfuscation, $Q_3$ denotes \emph{order quantiles} obfuscation with three quantiles, $B_3$ denotes \emph{fitness bucket} obfuscation with three buckets, $T_4$ denotes \emph{top individuals} obfuscation returning the top four individuals, and $A_{70}$ denotes \emph{above threshold} obfuscation returning the individuals with a fitness of at least 70~\% of the range between the best and worst actual fitness values.
For $Q_3$ obfuscation, the population is partitioned into three approximately equal-sized quantiles; due to the odd population size, the top quantile contains fewer solutions.
Although the solutions with indices 7 and 8 have identical fitness values, they are assigned to different quantiles, with ties broken arbitrarily.
For $B_3$ obfuscation, the fitness range is partitioned into three intervals of approximately equal length: $[67,100]$, $[34,67)$, and $[0,34)$.
The obfuscated fitness of a solution corresponds to the index of the interval containing the solution's fitness value.
For $T_4$ obfuscation, the four best solutions with respect to the actual fitness are selected; since the solutions with indices 4 and 5 are tied, the tie is broken arbitrarily.
For $A_{70}$ obfuscation, the five solutions with a fitness of at least 70 are selected.

\begin{table}[ht!]
    \centering
    \footnotesize
    \caption{Obfuscated and estimated fitness values using different obfuscation methods for a population of eleven solutions. The solutions of the population are sorted by the actual fitness values for illustration purposes. $O$: \emph{order}; $Q_3$: \emph{order quantiles} with $k=3$; $B_3$: \emph{fitness buckets} with $k=3$; $T_4$: \emph{top individuals} with $k=4$, $A_{70}$: \emph{above threshold} with $k=70$.}
    \begin{tabular}{rrcccccrrrrrr}
        \toprule
         \multicolumn{1}{c}{\multirow{2}{*}{Solution Index}} & \multicolumn{1}{c}{\multirow{2}{*}{Actual Fitness}} & \multicolumn{5}{c}{Obfuscated Fitness} & \multicolumn{5}{c}{Estimated Fitness} \\
         \cmidrule(lr){3-7}
         \cmidrule(lr){8-12}
         & & $O$ & $Q_{3}$ & $B_{3}$ & $T_{4}$ & $A_{70}$ & $O$ & $Q_{3}$ & $B_{3}$ & $T_{4}$ & $A_{70}$ \\
         \midrule
         1  & 100 &  0 & 0 & 0 & true  & true  &  100 &  100 &  100 &  100 &  100 \\
         2  &  90 &  1 & 0 & 0 & true  & true  &   80 &  100 &  100 &  100 &  100 \\
         3  &  80 &  2 & 0 & 0 & true  & true  &   60 &  100 &  100 &  100 &  100 \\
         4  &  70 &  3 & 1 & 0 & true  & true  &   40 &    0 &  100 &  100 &  100 \\
         5  &  70 &  4 & 1 & 0 & false & true  &   20 &    0 &  100 & -100 &  100 \\
         6  &  65 &  5 & 1 & 1 & false & false &    0 &    0 &    0 & -100 & -100 \\
         7  &  30 &  6 & 1 & 2 & false & false &  -20 &    0 & -100 & -100 & -100 \\
         8  &  30 &  7 & 2 & 2 & false & false &  -40 & -100 & -100 & -100 & -100 \\
         9  &  20 &  8 & 2 & 2 & false & false &  -60 & -100 & -100 & -100 & -100 \\
         10 &  10 &  9 & 2 & 2 & false & false &  -80 & -100 & -100 & -100 & -100 \\
         11 &   0 & 10 & 2 & 2 & false & false & -100 & -100 & -100 & -100 & -100 \\
         \bottomrule
    \end{tabular}
    \label{tab:obfuscation_estimation}
\end{table}

Obfuscation reduces the information that is revealed to an honest-but-curious platform provider, which we illustrate by estimating the number of equations and inequations that can be derived from obfuscated evaluation results during an optimization process.
Consider the previous example of a linear sum assignment with a $5\times5$ weight matrix $C=(c_{ij})$ and two candidate solutions $x$ and $y$ encoded as vectors (0, 1, 2, 3, 4) and (0, 2, 1, 3, 4).
When using \emph{order} obfuscation, the fitness of the best solution is revealed for each population.
Let $x$ and $y$ be the best solutions of different populations with a fitness of 100 and 90, respectively.
Based on this information, the provider learns
\[
  c_{1,1} + c_{2,2} = c_{1,2} + c_{2,1} + 10.
\]
Comparing the best solutions of 500 populations, the provider could construct up to
\[
  \binom{500}{2} = \numprint{124750}
\]
equations, assuming that no duplicate solutions are generated.
In addition, the order of solutions within a population is revealed to the provider.
Let $x$ and $y$ be in the same population, with $x$ having a better rank than $y$, the provider learns
\[
  c_{1,1} + c_{2,2} \geq c_{1,2} + c_{2,1},
\]
not knowing the precise difference between $x$ and $y$ in terms of fitness.
Comparing the ranked solutions within each population, the provider could construct up to
\[
  \binom{300}{2} \times 500 = \numprint{22450000}
\]
inequations after 500 iterations with a population size of 300, assuming that no duplicate solutions are generated.
The provider can additionally infer relations between the best solutions and solutions of other populations that have a smaller maximum fitness.
Suppose the populations are ordered by the fitness of their best solution, and the candidate solutions $x$ and $y$ are in different populations.
If solution $x$ is the best solution of a population that has a better rank than the population that contains $y$, the above inequation can also be constructed.
Inequations can be constructed for the best solution of the best-ranked population and the solutions of the remaining 499 populations, the best solution of the second-best-ranked population and the solutions of the remaining 498 populations, and so on.
For a population size of 300, the best solution of a population is compared with 299 solutions of each worse-ranked population.
There is no need to construct inequations for the best solutions of different populations, since equations already exist for these relations.
The provider could therefore construct up to
\[
  \sum_{i=1}^{500} 299 \times (500 - i) = \numprint{37300250}
\]
inequations after 500 iterations, assuming no duplicate solutions are generated.

The remaining obfuscation methods prevent constructing equations and further reduce the number of inequations.
In contrast to \emph{order} obfuscation, these obfuscation methods only reveal the order between subsets of solutions within a population.
Neither the identity of the best solution nor the order within subsets is revealed, except with \emph{fitness buckets} obfuscation, if only a single solution is in the best bucket.

When using \emph{order quantiles} obfuscation, only if the solutions $x$ and $y$ are in the same population and have a different quantile index, the provider can derive an inequation.
For example, using $Q_{10}$ with a population size of 300 results in 30 individuals being assigned to each quantile.
In this case, an inequation can be constructed for each solution of the best 30 and the worse 270, the second-best 30 and the worse 240, and so on.
The provider could therefore construct up to
\[
    (\sum_{i=1}^{10} 30 \times (300 - 30 \times i)) \times 500 = \numprint{20250000}
\]
inequations after 500 iterations, assuming no duplicate solutions are generated.

When using \emph{top individuals} obfuscation, only if the solutions $x$ and $y$ are in the same population and $x$ is a top solution while $y$ is not a top solution, the provider can derive an inequation.
For example, using $T_{30}$, the top 30 individuals are returned.
With a population size of 300, an inequation can be constructed for each solution of the top 30 and the worse 270.
The provider could therefore construct up to
\[
    30 \times (300-30) \times 500 = \numprint{4050000}
\]
inequations after 500 iterations, assuming no duplicate solutions are generated.

Estimating the number of inequations for \emph{fitness buckets} and \emph{above threshold} obfuscation is not as straightforward due to varying numbers of solutions in each bucket and above the threshold in each population.
When using \emph{fitness buckets} obfuscation, and only a single solution is in the best bucket in at least two iterations, then the provider can construct an equation.
For the remaining relations, the provider is limited to constructing only inequations.
The same issue occurs for \emph{above threshold} obfuscation but with the Privacy Engine implementation always returning at least three solutions, equations cannot be constructed.

The Privacy Engine may implement additional safeguards in case the platform provider attempts to learn more about the objective function than the above mentioned equations and inequations via systematically querying the Privacy Engine.
However, we assume an honest-but-curious platform provider who follows the protocol.
In addition, the log files of MPC nodes, which are hosted by a subset of participants, can reveal whether the platform provider deviates from the protocol and performs suspicious queries.

\subsection{Fitness Estimation}

The Optimizer estimates the fitness for obfuscated objectives based on the obfuscated evaluation result.
Table~\ref{tab:obfuscation_estimation} shows obfuscated evaluation results for a population of eleven solutions when using different obfuscation methods.
Our Optimizer implementation estimates the fitness for objective~$i$ of solutions in a population $P$ based on the revealed maximum fitness of the population $f_i^{\max} = \max\limits_{x \in P} f_i(x)$.
The Optimizer first estimates the minimum fitness of the population
\[
    f_i^{\min} = f_i^{\max} - 2 \times |f_i^{\max}|.
\]
Note that $f_i^{\min}$ is an estimate of the minimum fitness, i.e., $f_i^{\min} \neq \min\limits_{x \in P} f_i(x)$.
The Optimizer then calculates a distance between the estimated fitness of different solution subsets within the population
\[
    d^P = (f_i^{\max} - f_i^{\min}) / (u - 1)
\]
where $u$ is the number of distinct obfuscated fitness values. 
The value of $u$ corresponds to the population size for \emph{order} obfuscation, the number of buckets or quantiles for \emph{fitness buckets} and \emph{order quantiles} obfuscation, and two (true/false) for \emph{above threshold} and \emph{top individuals} obfuscation.
The estimated fitness $e^P_i$ of a solution $x$ is calculated based on its obfuscated fitness, the maximum fitness $f_i^{\max}$ and distance $d^P$ as follows:
\[
    e^P_i(x) = f_i^{\max} - m(r_i^P(x)) \times d^P
\]
where $m(r_i^P(x))$ is 0 if $r_i^P(x) = true$, 1 if $r_i^P(x) = false$, and otherwise $r_i^P(x)$.

Table~\ref{tab:obfuscation_estimation} shows the estimated fitness of solutions based on an obfuscated evaluation result for different obfuscation methods.
The maximum fitness of the population $f_i^{\max}$ is \numprint{100}, which results in an estimated minimum fitness $f_i^{\min}$ of \numprint{-100}.
The value of the distance $d^P$ is 20 for \emph{order} obfuscation, 100 for $Q_3$ and $B_3$ obfuscation, and 200 for $T_4$ and $A_{70}$ obfuscation.

The fitness can be estimated in a variety of ways.
There may be different options for estimating the minimum fitness and calculating the distance.
The distances between solution subsets may, for example, decrease exponentially rather than remain uniform.
This can have an impact on multi-objective optimization, where algorithms consider densities in the objective space.

\section{Experimental Setup}
\label{sec:setup}
In this section, we describe the experimental setup used to evaluate the proposed approach. 
In particular, we investigate the impact of obfuscation on the performance of evolutionary algorithms. 
To this end, we consider multiple problem classes and perform parameter tuning for each class to determine suitable configurations of the evolutionary algorithm. 
We then use these configurations to run the evolutionary algorithm under different obfuscation configurations.

We refer to online repositories providing implementations of the Optimizer~\cite{optimizer} and the Privacy Engine~\cite{privacy-engine}, as well as to an experimental suite~\cite{exp-suite} comprising scripts for instance generation, experiment execution, result analysis, and visualization. 
In addition, we refer to online resources~\cite{exp-suite-resources} that include the generated instances, experimental data, and analysis results.
The scripts in the experimental suite employ fixed random seeds to ensure reproducibility of the experiments~\cite{lopez2021}.

\subsection{Problem Instances}
For the experiments, we focus on three problem classes: balanced single-objective assignment problem (AP), symmetric single-objective traveling salesperson problem (TSP), and balanced multi-objective AP (MOAP) with two objectives.
The AP and MOAP problem instances were randomly generated, whereas the TSP instances were taken from TSPLIB~\cite{reinelt1991}\footnote{At the time of writing, the TSPLIB website was no longer accessible. Copies of the instances used are available in our online repository~\cite{exp-suite-resources}.}.
For all instances, the problem size was fixed at $m = n = 100$.
In particular, we used the following instances.

\begin{itemize}
 \item For the AP, 15 instances were generated, with utility values randomly drawn from the range [0, \numprint{1000}].
The optimal solutions for these instances were computed using a deterministic algorithm.

 \item For the MOAP, 15 instances with two objectives were generated, with utility values randomly drawn from the same range as for the AP.
 For each MOAP instance, a subset of Pareto-optimal solutions, which we refer to as \emph{ground truth}, was identified using a deterministic algorithm on 101 differently weighted combinations of the two objectives.
 For each $i \in \{0,...,100\}$, we created a weighting tuple $t_i = (\omega_{i,1}, \omega_{i,2})$ where $\omega_{i,1} = i \div 100$ and $\omega_{i,2} = 1-\omega_{i,1}$.
 To identify the ground truth, we normalized the values of each weight matrix and, for each tuple $t_i$, we multiplied $\omega_{i,1}$ with the first matrix and $\omega_{i,2}$ with the second matrix, summed the resulting matrices into a single matrix, and applied a deterministic algorithm.

 \item For the TSP, five instances with known optimal solutions were selected from TSPLIB. 
\end{itemize}

The problem instances were split into a set for parameter tuning and a set for performance evaluation.
For the AP and MOAP, ten instances were used for parameter tuning and five instances for performance evaluation.
For the TSP, three instances were used for parameter tuning and two instances for performance evaluation.

\subsection{Optimizer}
Our implementation of the Optimizer~\cite{optimizer} is a framework for executing evolutionary algorithms, which can be configured as a genetic algorithm for single-objective optimization and as NSGA-II~\cite{deb2002} for multi-objective optimization.
Algorithm~\ref{alg:evolutionary_algorithm} describes the framework.
Starting from an initial population of $\mu$ randomly generated individuals, the algorithm iteratively selects parents, generates offspring through recombination, applies mutation, evaluates the offspring, and, if necessary, reduces the population to $\mu$ survivor individuals.

\begin{algorithm}[h]
\small
\caption{Framework for evolutionary algorithms}
\label{alg:evolutionary_algorithm}
\begin{algorithmic}[1]
\State $t \gets 0$ \Comment{Initialize generation counter}
\State $I_{t} \gets \mu$ random individuals \Comment{Create initial population}
\State FitnessEvaluation($I_{t}$) \Comment{Evaluate individuals}
\State $P_{t} \gets$ SelectSurvivors($I_{t}, \mu$) \Comment{Select $\mu$ survivor individuals}
\While{termination criterion not met}
    \State $t \gets t + 1$ \Comment{Increment generation counter}
    \State $\Pi \gets $ SelectParents($P_{t-1}, \mu - \kappa + 1$) \Comment{Select enough parents for crossover}
    \State $O_{t} \gets $ Crossover($\Pi, \mu - \kappa$) \Comment{Create offspring}
    \State $O_{t} \gets$ Mutation($O_{t}$) \Comment{Mutate offspring}
    \State $I_{t} \gets O_{t} \cup $ SelectSurvivors($P_{t-1}, \kappa$) \Comment{Add $\kappa$ elite individuals for re-evaluation}
    \If{ReEvaluateParents} \Comment{Check if parent population should be re-evaluated}
        \State $I_{t} \gets I_{t} \cup P_{t-1}$ \Comment{Add parent population for re-evaluation}
    \EndIf
    \State FitnessEvaluation($I_{t}$) \Comment{Evaluate individuals}
    \State $P_{t} \gets $ SelectSurvivors($I_{t}, \mu$) \Comment{Select $\mu$ survivor individuals}
\EndWhile
\end{algorithmic}
\end{algorithm}

The framework supports different parent and survivor selection mechanisms.
For a genetic algorithm, the framework supports \emph{tournament selection} for parent selection, and a \emph{truncation selection} that selects the top individuals as survivors.
For NSGA-II, the framework supports \emph{binary tournament selection} based on Pareto rank and crowding distance for parent selection, and \emph{non-dominated sorting} and \emph{crowding distance sorting} to select the top individuals as survivors.
The number of parents depends on the number of elite individuals $\kappa$.
If the number of offspring individuals $\mu - \kappa$ is odd, selecting $\mu - \kappa + 1$ parents ensures a sufficient number of parents for crossover.

The framework for evolutionary algorithms can be configured with different crossover and mutation operators.
Crossover produces $\mu - \kappa$ offspring individuals by pairwise recombination of parents.
For each pair of parents, the crossover probability decides whether these individuals are recombined or added to the offspring without recombination.
The operators include \emph{cycle}, \emph{edge}, \emph{order}, \emph{partially mapped}, and \emph{uniform order-based} crossover, which are implemented based on Eiben and Smith~\shortcite{eiben2015-4} and Kruse et al.~\shortcite{kruse2022-12}.
For each offspring individual, a mutation probability decides whether the individual is mutated.
The operators include \emph{insert}, \emph{inversion}, \emph{scramble}, and \emph{swap} mutation which are implemented based on Eiben and Smith~\shortcite{eiben2015-4}.

The fitness evaluation passes a set of individuals to the evaluation component of the Optimizer.
For non-obfuscated objectives, the Optimizer calculates the fitness of the individuals.
For obfuscated objectives, the Optimizer sends the set of individuals to the Privacy Engine instance associated with the objective and estimates the fitness of the individuals after receiving an obfuscated evaluation result from the Privacy Engine.

The framework for evolutionary algorithms supports elitism by re-evaluating elite individuals together with the offspring.
If an objective is obfuscated, the estimated fitness of an individual depends on the characteristics of its population, such as the maximum fitness.
The estimated fitness of elite individuals is therefore not comparable to the estimated fitness of individuals in other generations.
The framework uses the mechanism for survivor selection to select $\kappa$ elite individuals from the parent population and combines these elite individuals with the offspring population for fitness evaluation.

The framework for evolutionary algorithms supports re-evaluating the parent population together with the offspring population.
NSGA-II ensures elitism by combining parent and offspring populations for non-dominated sorting~\cite{deb2002}.
If an objective is obfuscated, the estimated fitness between parent and offspring population are not comparable, which makes non-dominated sorting difficult.
The framework can therefore be configured to combine the parent population with the offspring population for fitness evaluation, which enables non-dominated sorting.

In the experiments, the termination criterion is 500 generations, i.e., $t > 500$.
The genetic algorithm is configured with a population size of $\mu = 300$, including $\kappa = 15$ elite individuals, while NSGA-II is configured with a population size of $\mu = 150$ and re-evaluation of the parent population.
These configurations result in evaluating 300 individuals for each generation where $t > 0$ for both evolutionary algorithms.
The Optimizer stores each population $P_t$ and selects the optimal or non-dominated individuals in $P_t$ as the optimization result of that generation.
The optimization results are used to calculate the performance metrics.

The experiments for parameter tuning were performed without obfuscation.
This study neither aims to identify an optimal configuration of evolutionary algorithms for each obfuscation configuration nor aims to compare different genetic operators or algorithms in combination with obfuscation.
Instead, for each problem class, we optimize a set of parameters for the configuration of an evolutionary algorithm, and use these parameters and different obfuscation configurations to investigate the impact of obfuscation on the performance of the evolutionary algorithm.

\subsection{Privacy Engine}
The experiments for performance evaluation were performed with a simulation of the Privacy Engine to reduce the runtime.
When simulating the Privacy Engine, the Optimizer calculates the actual fitness of individuals in a population, obfuscates the evaluation result, and estimates the fitness of the individuals based on the obfuscated results.
Despite the actual fitness being available, evolutionary algorithms only use the estimated fitness for obfuscated objectives.
The optimization results on the AP and MOAP when simulating the Privacy Engine are identical to those obtained when using our implementation of the Privacy Engine.
The fitness evaluation for the TSP is currently only supported in the simulation of the Privacy Engine.

We aim to compare the performance of evolutionary algorithms across obfuscation configurations.
Therefore, we must consider computational differences between obfuscation methods.
We measured the time required for evaluating a population of 300 individuals with our implementation of the Privacy Engine for the AP.
These results serve as the basis to also compare performance on the TSP and the MOAP.
The fitness calculations for the AP and TSP are similar, as both involve summing values from a matrix.
The fitness calculation for the MOAP is basically the same as for the AP, but with multiple objectives.
Nevertheless, we expect the same runtime for the AP and MOAP, assuming that each obfuscated objective is associated with its own Privacy Engine instance, which allows fitness for different obfuscated objectives to be calculated in parallel.

Our implementation of the Privacy Engine uses the MP-SPDZ framework, which provides protocols for benchmarking computational costs in different security models~\cite{keller2020}.
We deployed three MPC nodes and assumed an honest majority as the security model, which means that the majority of the MPC nodes, i.e., two nodes, behave honestly.
%
% This assumption allows using lightweight secret-sharing protocols with reduced preprocessing costs.
%
We selected the rep-field protocol~\cite{rep3} for measuring the time required to evaluate a population because it outperformed other protocols in sorting a list of 100 elements.

Table~\ref{tab:obfuscation_configs} reports for each obfuscation method the average time required to evaluate a population of 300 individuals on an AP instance with a problem size of 100×100 across 101 runs.
The runtimes were measured using MP-SPDZ at version 0.4.1, and Ubuntu 24.04.3 LTS on a 16 × Intel Xeon W-2245 CPU running at 3.90 GHz.
We did not consider network latency, which would have a strong impact on runtime.
Additional measurements, including latency, obtained from a separate execution of the benchmark scripts, are available online~\cite{privacy-engine}.
The configurations of the configurable obfuscation methods for these measurements were 5~buckets for \emph{fitness buckets} obfuscation, 75~\% threshold for \emph{above threshold} obfuscation, 5~quantiles for \emph{order quantiles} obfuscation, and 10 individuals for \emph{top individuals} obfuscation.

\begin{table}[]
    \centering
    \footnotesize
    \caption{The average evaluation time in seconds to evaluate a population of 300 individuals with an instance of our Privacy Engine implementation, the comparison generation, and the configurations for each obfuscation method, including no obfuscation, used in the performance experiments.}
    \label{tab:obfuscation_configs}
    \begin{tabular}{cccrrrc}
        \toprule
        Obfuscation Method & Average Time & Generation & \multicolumn{4}{c}{Configurations}  \\
        \midrule
        No Obfuscation & 0.27 & 500 & \multicolumn{4}{c}{-}\\
        Fitness Buckets & 2.45 & 54 & 5 & 10 & 20 & buckets \\
        Above Threshold & 0.74 & 181 & 80 & 90 & 95 & threshold (\%) \\
        Order & 0.78 & 172 & \multicolumn{4}{c}{-}\\
        Order Quantiles & 1.22 & 109 & 5 & 10 & 20 & quantiles \\
        Top Individuals & 0.79 & 170 & 60 & 30 & 15 & individuals\\
        \bottomrule
    \end{tabular}
\end{table}

To make the performance of evolutionary algorithms comparable across obfuscation methods, we introduce a time budget on the time available for evaluation.
The time budget is calculated from the average evaluation time per population without obfuscation with the Privacy Engine.
The average evaluation time is multiplied by 501 to include the initial population and 500 subsequent generations, which results in a time budget of 135.27 seconds.
For the sake of simplicity, we assume that the runtime of an evolutionary algorithm remains constant across generations and obfuscation configurations.
Table~\ref{tab:obfuscation_configs} reports for each obfuscation method the final generation that can be evaluated within the time budget under this assumption.
We only use the optimization results of these final generations for performance comparisons across obfuscation methods.

We selected three configurations for each configurable obfuscation method for the performance experiments.
These configurations are shown in Table~\ref{tab:obfuscation_configs}.
The configurations for \emph{fitness buckets} obfuscation are $B_5, B_{10},$ and $B_{20}$.
When using $B_5$, the fitness range covered by each bucket is $100~\% \div 5 = 20~\%$.
The fitness range covered by each bucket decreases to 10~\% when using $B_{10}$ and further decreases to 5~\% when using $B_{20}$.
The configurations for \emph{above threshold} obfuscation are 100~\% minus the fitness range covered by each bucket for the different configurations of \emph{fitness buckets} obfuscation.
The configurations for \emph{order quantiles} obfuscation are $Q_5, Q_{10},$ and $Q_{20}$.
When using $Q_5$ and evaluating 300 individuals, $300 \div 5 = 60$ individuals are assigned to each quantile.
The number of individuals assigned to each quantile decreases to 30 when using $Q_{10}$ and further decreases to 15 when using $Q_{20}$.
The configurations for \emph{top individuals} obfuscation are equal to the number of individuals that are assigned to each quantile for the different configurations of \emph{order quantiles} obfuscation.

\subsection{Metrics}
We distinguish between measuring the performance of evolutionary algorithms for parameter tuning and for performance evaluation.
Parameter tuning is performed without obfuscation and performance is aggregated across instances, while performance evaluation is performed with obfuscation and performance is not aggregated across instances.

In the experiments for parameter tuning of the genetic algorithm on the AP and the TSP, we compute a relative error $\epsilon$ for each run.
Smaller values for the relative error $\epsilon$ indicate better solutions, whereas $\epsilon = 0$ corresponds to the optimal solution.
The relative error $\epsilon$ is calculated for a solution~$x$ based on its fitness $f(x)$ and the optimal fitness $f^\star$ of an instance as follows:
\[
\epsilon(x) = (f^\star - f(x)) \div f^\star \text{ for the AP,} \quad \epsilon(x) = (f(x) - f^\star) \div f^\star \text{ for the TSP.}
\]

For the MOAP, we base our evaluation on the \emph{generational distance} (GD) and the \emph{inverted GD} (IGD) metrics.
To calculate these metrics, we use the previously identified subset of Pareto-optimal solutions, which we refer to as \emph{ground truth}.
The GD corresponds to the average distance from each non-dominated solution found to its nearest solution in the ground truth.
The IGD corresponds to the average distance from each solution in the ground truth to its nearest non-dominated solution found.
Smaller values for GD or IGD indicate better solution sets.
Ishibuchi et al.~\shortcite{ishibuchi2015} propose a modified distance calculation that additionally takes into account the dominance relation between two solutions.
Using the modified distance calculation to compute GD and IGD, we adopt the notions GD\textsuperscript{+} and IGD\textsuperscript{+} from Ishibuchi et al.~\shortcite{ishibuchi2015} for these metrics.

In the experiments for parameter tuning of NSGA-II on the MOAP, we normalized the fitness values using the ideal and nadir points from the ground truth before calculating GD\textsuperscript{+} and IGD\textsuperscript{+}.
Smaller values for the normalized fitness $\tilde{f}_{i}$ indicate better solutions with regard to objective~$i$, whereas $\tilde{f}_{i} = 0$ and $\tilde{f}_{i} = 1$ correspond to the ideal and nadir fitness for objective~$i$ in the ground truth.
The normalized fitness $\tilde{f}_{i}$ of a solution $x$ for objective $i$ is calculated based on its fitness $f_i(x)$, the ideal fitness $f_i^\star$ and the nadir fitness $f_i^{nad}$ for objective $i$ in the ground truth as follows:
\[
\tilde{f}_{i}(x) = (f_i^\star - f_i(x)) \div (f_i^\star - f_i^{nad})
\]

In the performance experiments, we calculated the metrics using the set of \emph{estimated} optimal or non-dominated individuals within a population.
This set consists of individuals that are optimal or Pareto-optimal based on the estimated fitness for obfuscated objectives and actual fitness for non-obfuscated objectives.
This set may thus include individuals that are suboptimal or dominated when considering the actual fitness for obfuscated objectives.
We include these individuals when calculating the metrics to better reflect the outcome from a decision-maker's perspective, as only estimated fitness values are available for the obfuscated objectives.

We measure the performance of the genetic algorithm on the AP and the TSP using the mean fitness of the set of estimated optimal individuals and the performance of NSGA-II on the MOAP using GD\textsuperscript{+} and IGD\textsuperscript{+} of the set of estimated non-dominated individuals.
Higher values for the mean fitness indicate better solution sets for the AP, whereas smaller values indicate better solution sets for the TSP.
Smaller values for GD\textsuperscript{+} and IGD\textsuperscript{+} indicate better solution sets for the MOAP.

\subsection{Experiments}
The experiments for parameter tuning were performed with the parameter optimization framework Optuna~\cite{optuna} for each problem class.
Optuna takes as input an objective function to evaluate a parameter set.
Our objective function configures an evolutionary algorithm and executes the configured algorithm three times on each of the instances selected for parameter tuning.
For each instance, the median performance of Generation~500 over three independent runs was selected.
The performance of a parameter set was determined by the mean of the selected median performance across instances.

Optuna was configured to sample and evaluate a total of 300 parameter sets for each problem class.
Before using the multivariate tree-structured Parzen estimator for sampling parameter sets, 50 parameter sets were randomly sampled. 
Table~\ref{tab:ap_config}, Table~\ref{tab:tsp_config}, and Table~\ref{tab:moap_config} report the options from which the parameter values for the configuration of the evolutionary algorithms were sampled and the best parameter sets found for the AP, TSP and MOAP, respectively.
The parameter set for the AP achieved a mean median relative error of \numprint{0.019}, and the parameter set for the TSP achieved \numprint{0.043}.
For the MOAP, two parameter sets, which were non-dominated with respect to GD\textsuperscript{+} and IGD\textsuperscript{+}, were found:  
Set~1 achieved a mean median GD\textsuperscript{+} of \numprint{0.116} and a mean median IGD\textsuperscript{+} of \numprint{0.126}, while Set~2 achieved \numprint{0.113} and \numprint{0.127}.
We used Set~2 in the performance experiments due to having a smaller sum of GD\textsuperscript{+} and IGD\textsuperscript{+} than Set~1.

\begin{table}
    \centering
    \footnotesize
    \caption{Options for configuring the genetic algorithm using tournament selection for the assignment problem and the best parameter set found.}
    \label{tab:ap_config}
    \begin{tabular}{lcc}
        \toprule
        \multicolumn{1}{c}{Parameter} & Configuration Options & Best Parameter Set\\
        \midrule
        Number of Generations & 500 & 500 \\
        Population Size & 300 & 300 \\
        Number of Elite Individuals & 15 & 15\\
        Tournament Size & \{2, 4, 6, 8, 10, 12, 14, 16, 18, 20\} & 18 \\
        Crossover Type & \{Partially Mapped, Uniform Order-Based, Cycle\} & Cycle \\
        Crossover Probability & \{60, 65, 70, 75, 80, 85, 90, 95, 100\}~\% & 95~\% \\
        Mutation Type & \{Swap, Insert, Scramble, Inversion\} & Swap \\
        Mutation Probability & \{60, 65, 70, 75, 80, 85, 90, 95, 100\}~\% & 100~\% \\
        \bottomrule
    \end{tabular}
\end{table}

\begin{table}
    \centering
    \footnotesize
    \caption{Options for configuring the genetic algorithm using tournament selection for the traveling salesperson problem and the best parameter set found.}
    \label{tab:tsp_config}
    \begin{tabular}{lcc}
        \toprule
        \multicolumn{1}{c}{Parameter} & Configuration Options & Best Parameter Set\\
        \midrule
        Number of Generations & 500 & 500 \\
        Population Size & 300 & 300 \\
        Number of Elite Individuals & 15 & 15\\
        Tournament Size & \{2, 4, 6, 8, 10, 12, 14, 16, 18, 20\} & 20 \\
        Crossover Type & \{Partially Mapped, Order, Edge\} & Edge \\
        Crossover Probability & \{60, 65, 70, 75, 80, 85, 90, 95, 100\}~\% & 75~\% \\
        Mutation Type & \{Swap, Insert, Scramble, Inversion\} & Inversion \\
        Mutation Probability & \{60, 65, 70, 75, 80, 85, 90, 95, 100\}~\% & 100~\% \\
        \bottomrule
    \end{tabular}
\end{table}

\begin{table}
    \centering
    \footnotesize
    \caption{Options for configuring NSGA-II for the multi-objective assignment problem and the best parameter sets found.}
    \label{tab:moap_config}
    \begin{tabular}{lccc}
        \toprule
        \multicolumn{1}{c}{\multirow{2}{*}{Parameter}} & \multirow{2}{*}{Configuration Options} & \multicolumn{2}{c}{Best Parameter Sets}\\
        \cmidrule(lr){3-4}
        & & Set 1 & Set 2 \\
        \midrule
        Number of Generations & 500 & 500 & 500 \\
        Population Size & 150 & 150 & 150 \\
        Crossover Type & \{Partially Mapped, Uniform Order-Based, Cycle\} & Cycle & Cycle  \\
        Crossover Probability & \{60, 65, 70, 75, 80, 85, 90, 95, 100\}~\% & 100~\% & 100~\% \\
        Mutation Type & \{Swap, Insert, Scramble, Inversion\} & Swap & Swap \\
        Mutation Probability & \{60, 65, 70, 75, 80, 85, 90, 95, 100\}~\% & 80~\% & 75~\%\\
        \bottomrule
    \end{tabular}
\end{table}

The performance experiments were conducted by configuring the evolutionary algorithms with the parameter sets from the Tables~\ref{tab:ap_config}, \ref{tab:tsp_config}, and \ref{tab:moap_config}, depending on the problem class, and executing 31 runs for each obfuscation configuration in Table~\ref{tab:obfuscation_configs} on the problem instances selected for performance evaluation.
For the MOAP, we consider two different scenarios: one in which only one objective is obfuscated, and another in which both objectives are obfuscated.

We use the same set of random seeds $s_r \in \{0,...,30\}$ for configuring the runs of the evolutionary algorithms across obfuscation configurations.
More precisely, the first run with each obfuscation configuration on a problem instance is configured with $s_r = 0$, the second run with $s_r=1$, and so forth.
The runs with the same random seed thus start from the same initial population and allow for a pairwise comparison of the performance across obfuscation configurations.

\section{Experimental Results}
\label{sec:results}

In this section, we present the results of the performance experiments, demonstrating the impact of different obfuscation configurations on the performance of evolutionary algorithms using different problem classes.
Since the results are similar across instances, for each problem class, we present only the results of the alphabetically first instance; the results for the other instances are available online~\cite{exp-suite-resources}.
For each obfuscation configuration, we visualize convergence over 500 generations and report the median and interquartile range (IQR) for both the initial population and Generation~500.
The performance of the initial population is reported to indicate the performance of random solutions.

We also compare the performance of evolutionary algorithms across obfuscation configurations using the comparison generations (see Table~\ref{tab:obfuscation_configs}), i.e., the final generations that can be evaluated within a time budget.
The time budget corresponds to the time required by the Privacy Engine to evaluate 300 individuals on an AP instance over 501 iterations without obfuscation.
When comparing the performance of runs with obfuscation configuration~$X$ to the runs with obfuscation configuration~$Y$, the runs with $X$ and $Y$ that were executed with the same random seed $s_r$ were paired.
We assessed statistical significance using two-sided Wilcoxon signed-rank tests with a significance level of $\alpha = 0.05$.

The matched-pairs rank-biserial correlation is reported as an effect size for the comparison.
We rank the performance differences between the runs with the same random seed, and calculate the matched-pairs rank-biserial correlation as the difference between the proportion of the favorable rank sum and the proportion of the unfavorable rank sum~\cite{kerby2014}.
The value range of the rank-biserial correlation is [-1, 1].
A value of 1 indicates that, for all pairs of runs with configuration $X$ and $Y$, runs with $X$ achieved a higher performance.
The rank-biserial correlation is antisymmetric: if the value is 0.5 when comparing the runs with~$X$ to the runs with~$Y$, then the value is –0.5 when comparing the runs with~$Y$ to the runs with~$X$.

The interpretation of the matched-pairs rank-biserial correlation differs between the problem classes.
For the AP, positive rank-biserial correlation values indicate better performance, as higher values of mean fitness correspond to better solution sets.
For the TSP and MOAP, negative values indicate better performance, as lower values of mean fitness (TSP) and GD\textsuperscript{+} and IGD\textsuperscript{+} (MOAP) correspond to better solution sets.

Performance comparisons are summarized by counting the number of wins for each obfuscation configuration.
A configuration $X$ wins against $Y$ if the difference in performance between $X$ and $Y$ is statistically significant and the matched-pairs rank-biserial correlation indicates that $X$ achieves a better performance.
This summary provides insights into which obfuscation configurations result in better performance than others, rather than indicating which configurations perform well.

\subsection{Assignment Problem}

\begin{figure}
    \centering
    \includegraphics[width=\linewidth]{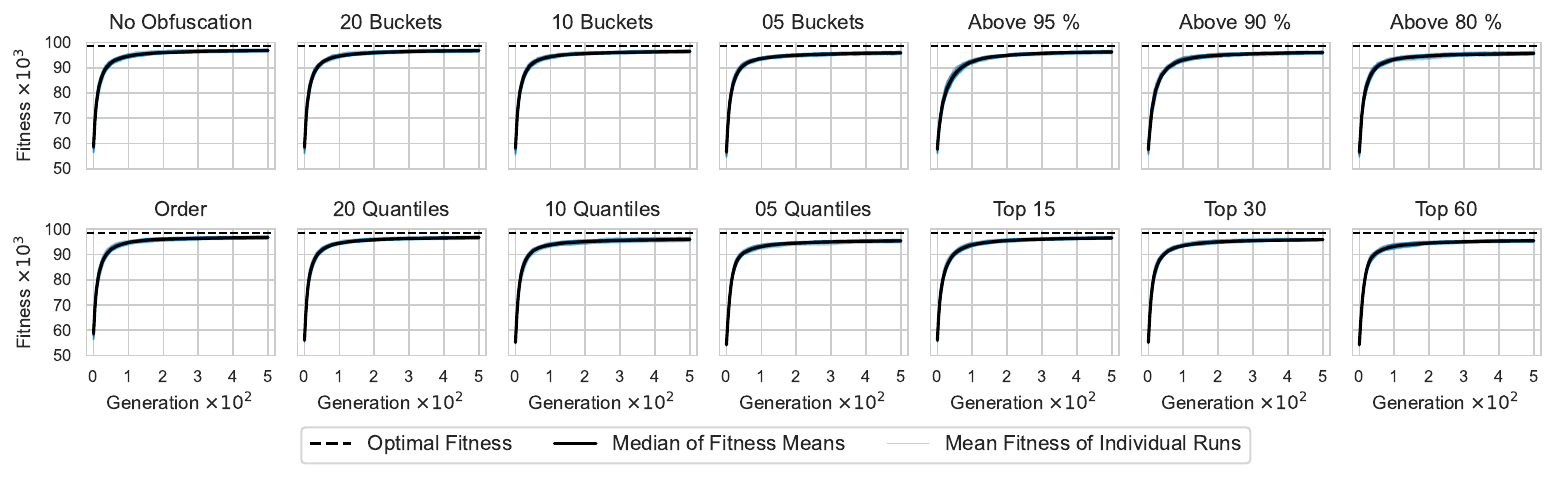}
    \caption{Median of fitness means per generation and mean fitness of 31 individual runs over 500 generations for each obfuscation configuration on instance \emph{ap1K100x100} with an optimal fitness of \numprint{98680}.}
    \label{fig:ap2_ap1K100x100_convergence}
\end{figure}

We present the results of a genetic algorithm on Instance \emph{ap1K100x100} with an optimal fitness of \numprint{98680}.
Figure~\ref{fig:ap2_ap1K100x100_convergence} shows the median of fitness means per generation and the mean fitness of 31 individual runs over 500 generations for each obfuscation configuration.
Individual runs of each obfuscation configuration closely follow the median curve, indicating low variability across runs.

Table~\ref{tab:ap1K100x100_kroD100_summary} reports, in the left-hand half, the median and IQR of the mean fitness across 31 runs for the initial population and Generation~500 for each obfuscation configuration.
The median of fitness means ranges from 96.65~\% of the optimum for $Q_5$ to 98.03~\% for $Q_{20}$ for the configurable obfuscation methods.
Within each obfuscation method, the median of fitness means of Generation~500 tends to improve, i.e., increase, when the number of buckets or quantiles is increased, the threshold is raised, or the number of top individuals is reduced.

The median and IQR of the mean fitness for the initial population in Table~\ref{tab:ap1K100x100_kroD100_summary} differ between obfuscation configurations, although the runs start from the same set of 31 initial populations.
This is because performance is measured using the set of estimated optimal solutions, which can be different between obfuscation configurations.
For \emph{order quantiles} and \emph{top individuals} obfuscation, the initial population has the same performance for the pairs $Q_{20}$ and $T_{15}$, $Q_{10}$ and $T_{30}$, and $Q_5$ and $T_{60}$ because the configurations of \emph{top individuals} obfuscation correspond to the number of individuals assigned to each quantile for the configurations of \emph{order quantiles} obfuscation.
This characteristic is not guaranteed for the configurations of \emph{fitness buckets} and \emph{above threshold} obfuscation because our implementation of \emph{above threshold} obfuscation returns at least three individuals.

\begin{table}
    \caption{Median and interquartile range (IQR) of the mean fitness, rounded to integers, for the initial population and Generation 500 across 31 runs for each obfuscation configuration on instances \emph{ap1K100x100} with an optimal fitness of \numprint{98680} and \emph{kroD100} with an optimal fitness of \numprint{21294}.}
    \footnotesize
    \label{tab:ap1K100x100_kroD100_summary}
    \begin{tabular}{rrrrrrrrr}
        \toprule
         \multicolumn{1}{c}{\multirow{3}{*}{Obfuscation}} & \multicolumn{4}{c}{ap1K100x100} & \multicolumn{4}{c}{kroD100} \\
         \cmidrule(lr){2-5}
         \cmidrule(lr){6-9}
         & \multicolumn{2}{c}{Initial Population} & \multicolumn{2}{c}{Generation 500} & \multicolumn{2}{c}{Initial Population} &  \multicolumn{2}{c}{Generation 500} \\
         \cmidrule(lr){2-3}
         \cmidrule(lr){4-5}
         \cmidrule(lr){6-7}
         \cmidrule(lr){8-9}
         & \multicolumn{1}{c}{Median} & \multicolumn{1}{c}{IQR}& \multicolumn{1}{c}{Median} & \multicolumn{1}{c}{IQR}& \multicolumn{1}{c}{Median} & \multicolumn{1}{c}{IQR}& \multicolumn{1}{c}{Median} & \multicolumn{1}{c}{IQR} \\
         \midrule
No Obfuscation & \numprint{58498} & \numprint{1766} & \numprint{96814} & \numprint{362} & \numprint{141188} & \numprint{4943} & \numprint{22579} & \numprint{948} \\
20 Buckets & \numprint{58307} & \numprint{1955} & \numprint{96725} & \numprint{310} & \numprint{141438} & \numprint{4909} & \numprint{23645} & \numprint{893} \\
10 Buckets & \numprint{57808} & \numprint{2192} & \numprint{96457} & \numprint{335} & \numprint{143072} & \numprint{6370} & \numprint{24987} & \numprint{755} \\
5 Buckets & \numprint{56417} & \numprint{1345} & \numprint{95834} & \numprint{334} & \numprint{146100} & \numprint{4709} & \numprint{31232} & \numprint{1657} \\
Above 95 \% & \numprint{57715} & \numprint{1037} & \numprint{96253} & \numprint{258} & \numprint{142895} & \numprint{2934} & \numprint{24523} & \numprint{1050} \\
Above 90 \% & \numprint{57707} & \numprint{912} & \numprint{96025} & \numprint{373} & \numprint{143463} & \numprint{3789} & \numprint{26324} & \numprint{1677} \\
Above 80 \% & \numprint{56417} & \numprint{1345} & \numprint{95643} & \numprint{365} & \numprint{146430} & \numprint{4709} & \numprint{31502} & \numprint{2392} \\
Order & \numprint{58498} & \numprint{1766} & \numprint{96750} & \numprint{256} & \numprint{141188} & \numprint{4943} & \numprint{22513} & \numprint{661} \\
20 Quantiles & \numprint{56183} & \numprint{399} & \numprint{96740} & \numprint{286} & \numprint{147369} & \numprint{954} & \numprint{22753} & \numprint{580} \\
10 Quantiles & \numprint{55310} & \numprint{348} & \numprint{96003} & \numprint{470} & \numprint{149719} & \numprint{798} & \numprint{25002} & \numprint{959} \\
5 Quantiles & \numprint{54331} & \numprint{336} & \numprint{95377} & \numprint{368} & \numprint{152428} & \numprint{826} & \numprint{30953} & \numprint{2325} \\
Top 15 & \numprint{56183} & \numprint{399} & \numprint{96567} & \numprint{351} & \numprint{147369} & \numprint{954} & \numprint{22935} & \numprint{1016} \\
Top 30 & \numprint{55310} & \numprint{348} & \numprint{95885} & \numprint{254} & \numprint{149719} & \numprint{798} & \numprint{25498} & \numprint{1272} \\
Top 60 & \numprint{54331} & \numprint{336} & \numprint{95510} & \numprint{370} & \numprint{152428} & \numprint{826} & \numprint{31318} & \numprint{2621} \\
         \bottomrule
    \end{tabular}
\end{table}

Table~\ref{tab:ap2_ap1K100x100_comparison} reports the median and IQR of the mean fitness of the comparison generations, together with the results from comparing the mean fitness of runs across obfuscation configurations.
The number of wins for each configuration is as follows: without obfuscation (13 wins), $O$ (12 wins), $T_{15}$ (11 wins), $A_{90}$ (7 wins), $Q_{20}$ (7 wins), $T_{30}$ (7 wins), $A_{95}$ (6 wins), $A_{80}$ (5 wins), $T_{60}$ (5 wins), $Q_{10}$ (4 wins), $Q_5$ (3 wins), $B_{20}$ (2 wins), $B_{10}$ (1 win), and $B_5$ (0 wins).
The runs with $T_{15}$ perform best among the configurable obfuscation methods.
The runs with \emph{fitness buckets} obfuscation suffer from the strongest impact on runtime and perform worst.

\begin{table}
    \centering
    \caption{Median and interquartile range (IQR) of the mean fitness of the comparison generation across 31 runs for each obfuscation configuration on Instance \emph{ap1K100x100} with an optimal fitness of \numprint{98680}. Matched-pairs rank-biserial correlations, rounded to one decimal place, are reported for comparing the performance of the configuration in each row with that in each column. Grey cells indicate statistically significant differences in performance. A dash (-) denotes no obfuscation.}
    \label{tab:ap2_ap1K100x100_comparison}
    \footnotesize
    \begin{tabular}{lrrrrrrrrrrrrrrr}
    \toprule
& \multirow{2}{*}{Median} & \multirow{2}{*}{IQR} & \multicolumn{13}{c}{Matched-Pairs Rank-Biserial Correlation} \\
\cmidrule(lr){4-16}
&        &     & - & $B_{20}$ & $B_{10}$ & $B_{5}$ & $A_{95}$ & $A_{90}$ & $A_{80}$ & $O$ & $Q_{20}$ & $Q_{10}$ & $Q_{5}$ & $T_{15}$ & $T_{30}$ \\
\midrule
- & \numprint{96814} & \numprint{362} \\
$B_{20}$ & \numprint{92354} & \numprint{345} & \cellcolor[HTML]{D3D3D3} -\numprint{1.0} \\
$B_{10}$ & \numprint{92075} & \numprint{774} & \cellcolor[HTML]{D3D3D3} -\numprint{1.0} & \cellcolor[HTML]{D3D3D3} -\numprint{0.4} \\
$B_{5}$ & \numprint{91244} & \numprint{689} & \cellcolor[HTML]{D3D3D3} -\numprint{1.0} & \cellcolor[HTML]{D3D3D3} -\numprint{1.0} & \cellcolor[HTML]{D3D3D3} -\numprint{0.9} \\
$A_{95}$ & \numprint{94739} & \numprint{446} & \cellcolor[HTML]{D3D3D3} -\numprint{1.0} & \cellcolor[HTML]{D3D3D3} \numprint{1.0} & \cellcolor[HTML]{D3D3D3} \numprint{1.0} & \cellcolor[HTML]{D3D3D3} \numprint{1.0} \\
$A_{90}$ & \numprint{94819} & \numprint{516} & \cellcolor[HTML]{D3D3D3} -\numprint{1.0} & \cellcolor[HTML]{D3D3D3} \numprint{1.0} & \cellcolor[HTML]{D3D3D3} \numprint{1.0} & \cellcolor[HTML]{D3D3D3} \numprint{1.0} & \numprint{0.2} \\
$A_{80}$ & \numprint{94621} & \numprint{511} & \cellcolor[HTML]{D3D3D3} -\numprint{1.0} & \cellcolor[HTML]{D3D3D3} \numprint{1.0} & \cellcolor[HTML]{D3D3D3} \numprint{1.0} & \cellcolor[HTML]{D3D3D3} \numprint{1.0} & -\numprint{0.4} & \cellcolor[HTML]{D3D3D3} -\numprint{0.6} \\
$O$ & \numprint{95931} & \numprint{593} & \cellcolor[HTML]{D3D3D3} -\numprint{1.0} & \cellcolor[HTML]{D3D3D3} \numprint{1.0} & \cellcolor[HTML]{D3D3D3} \numprint{1.0} & \cellcolor[HTML]{D3D3D3} \numprint{1.0} & \cellcolor[HTML]{D3D3D3} \numprint{1.0} & \cellcolor[HTML]{D3D3D3} \numprint{1.0} & \cellcolor[HTML]{D3D3D3} \numprint{1.0} \\
$Q_{20}$ & \numprint{94655} & \numprint{363} & \cellcolor[HTML]{D3D3D3} -\numprint{1.0} & \cellcolor[HTML]{D3D3D3} \numprint{1.0} & \cellcolor[HTML]{D3D3D3} \numprint{1.0} & \cellcolor[HTML]{D3D3D3} \numprint{1.0} & \numprint{0.3} & -\numprint{0.1} & \cellcolor[HTML]{D3D3D3} \numprint{0.5} & \cellcolor[HTML]{D3D3D3} -\numprint{1.0} \\
$Q_{10}$ & \numprint{94041} & \numprint{541} & \cellcolor[HTML]{D3D3D3} -\numprint{1.0} & \cellcolor[HTML]{D3D3D3} \numprint{1.0} & \cellcolor[HTML]{D3D3D3} \numprint{1.0} & \cellcolor[HTML]{D3D3D3} \numprint{1.0} & \cellcolor[HTML]{D3D3D3} -\numprint{0.9} & \cellcolor[HTML]{D3D3D3} -\numprint{0.9} & \cellcolor[HTML]{D3D3D3} -\numprint{0.8} & \cellcolor[HTML]{D3D3D3} -\numprint{1.0} & \cellcolor[HTML]{D3D3D3} -\numprint{1.0} \\
$Q_{5}$ & \numprint{93418} & \numprint{528} & \cellcolor[HTML]{D3D3D3} -\numprint{1.0} & \cellcolor[HTML]{D3D3D3} \numprint{1.0} & \cellcolor[HTML]{D3D3D3} \numprint{1.0} & \cellcolor[HTML]{D3D3D3} \numprint{1.0} & \cellcolor[HTML]{D3D3D3} -\numprint{1.0} & \cellcolor[HTML]{D3D3D3} -\numprint{1.0} & \cellcolor[HTML]{D3D3D3} -\numprint{1.0} & \cellcolor[HTML]{D3D3D3} -\numprint{1.0} & \cellcolor[HTML]{D3D3D3} -\numprint{1.0} & \cellcolor[HTML]{D3D3D3} -\numprint{0.9} \\
$T_{15}$ & \numprint{95205} & \numprint{446} & \cellcolor[HTML]{D3D3D3} -\numprint{1.0} & \cellcolor[HTML]{D3D3D3} \numprint{1.0} & \cellcolor[HTML]{D3D3D3} \numprint{1.0} & \cellcolor[HTML]{D3D3D3} \numprint{1.0} & \cellcolor[HTML]{D3D3D3} \numprint{1.0} & \cellcolor[HTML]{D3D3D3} \numprint{0.9} & \cellcolor[HTML]{D3D3D3} \numprint{1.0} & \cellcolor[HTML]{D3D3D3} -\numprint{1.0} & \cellcolor[HTML]{D3D3D3} \numprint{0.8} & \cellcolor[HTML]{D3D3D3} \numprint{1.0} & \cellcolor[HTML]{D3D3D3} \numprint{1.0} \\
$T_{30}$ & \numprint{94754} & \numprint{530} & \cellcolor[HTML]{D3D3D3} -\numprint{1.0} & \cellcolor[HTML]{D3D3D3} \numprint{1.0} & \cellcolor[HTML]{D3D3D3} \numprint{1.0} & \cellcolor[HTML]{D3D3D3} \numprint{1.0} & \numprint{0.2} & \numprint{0.0} & \cellcolor[HTML]{D3D3D3} \numprint{0.5} & \cellcolor[HTML]{D3D3D3} -\numprint{1.0} & \numprint{0.0} & \cellcolor[HTML]{D3D3D3} \numprint{1.0} & \cellcolor[HTML]{D3D3D3} \numprint{1.0} & \cellcolor[HTML]{D3D3D3} -\numprint{0.8} \\
$T_{60}$ & \numprint{94355} & \numprint{655} & \cellcolor[HTML]{D3D3D3} -\numprint{1.0} & \cellcolor[HTML]{D3D3D3} \numprint{1.0} & \cellcolor[HTML]{D3D3D3} \numprint{1.0} & \cellcolor[HTML]{D3D3D3} \numprint{1.0} & \cellcolor[HTML]{D3D3D3} -\numprint{0.6} & \cellcolor[HTML]{D3D3D3} -\numprint{0.7} & -\numprint{0.3} & \cellcolor[HTML]{D3D3D3} -\numprint{1.0} & \cellcolor[HTML]{D3D3D3} -\numprint{0.7} & \cellcolor[HTML]{D3D3D3} \numprint{0.5} & \cellcolor[HTML]{D3D3D3} \numprint{1.0} & \cellcolor[HTML]{D3D3D3} -\numprint{1.0} & \cellcolor[HTML]{D3D3D3} -\numprint{0.6} \\
    \bottomrule
    \end{tabular}
\end{table}

The median performance of Generation~500 across 31 runs is close to the optimum for all obfuscation configurations, with IQR indicating only minor variability.
Although reducing information leakage, the results show that obfuscation indeed negatively impacts performance when imposing the time budged on evaluation.
The binary fitness information provided by $T_{15}$ is sufficient for the genetic algorithm to find solution sets with a median of fitness means of 96.48~\% of the optimum within the time budget.

\subsection{Traveling Salesperson Problem}
\begin{figure}
    \centering
    \includegraphics[width=\linewidth]{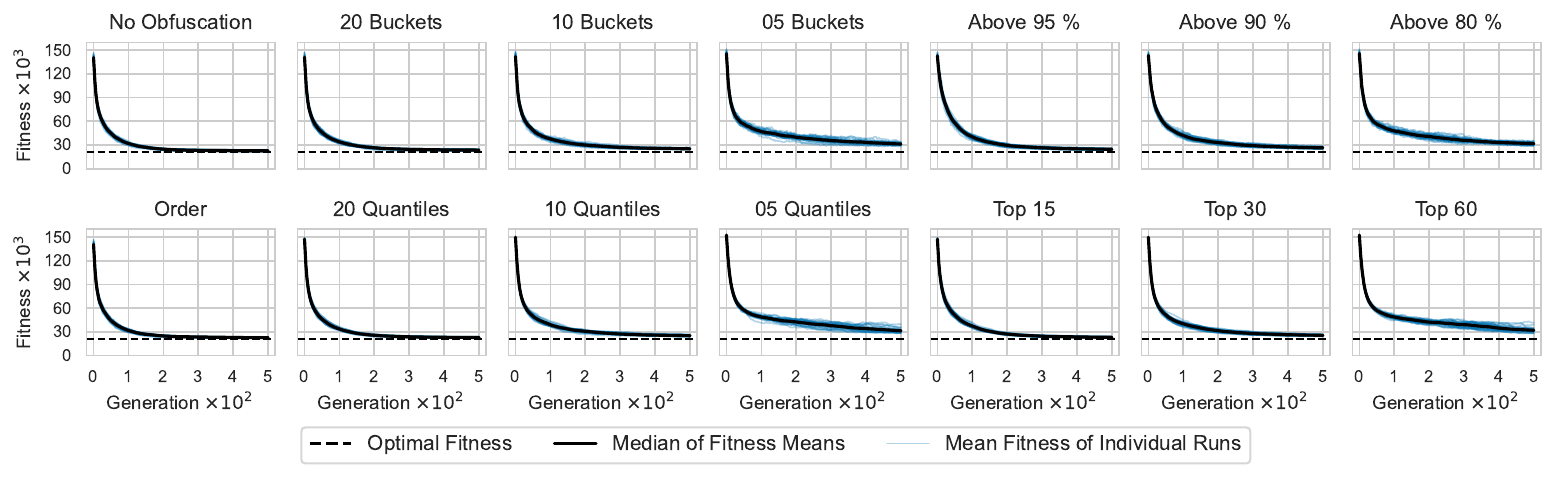}
    \caption{Median of fitness means per generation and mean fitness of 31 individual runs over 500 generations for each obfuscation configuration on Instance \emph{kroD100} with an optimal fitness of \numprint{21294}.}
    \label{fig:tsp2_kroD100_convergence}
\end{figure}

We present the results of a genetic algorithm on instance \emph{kroD100} with an optimal fitness of \numprint{21294}.
Figure~\ref{fig:tsp2_kroD100_convergence} shows the median of fitness means per generation and the mean fitness of 31 individual runs over 500 generations for each obfuscation configuration.
Individual runs with $B_5$, $Q_5$, $A_{80}$, and $T_{60}$ deviate slightly from the median curve after the initial fitness decrease.
Individual runs with the remaining obfuscation configurations closely follow the median curve, indicating low variability across runs.

Table~\ref{tab:ap1K100x100_kroD100_summary} reports, in the right-hand half, the median and IQR of the mean fitness across 31 runs for the initial population and Generation~500 for each obfuscation configuration.
The median of fitness means ranges from 106.85~\% of the optimum for $Q_{20}$ to 147.94~\% for $A_{80}$ for the configurable obfuscation methods.
We can observe the same pattern as for the AP; within each obfuscation method, the median of fitness means of Generation~500 tends to improve, i.e., decrease, when the number of buckets or quantiles is increased, the threshold is raised, or the number of top individuals is reduced.
For configurations $B_5$, $Q_5$, $A_{80}$ and $T_{60}$, the deviations from the optimum are quite considerable.

Table~\ref{tab:tsp2_kroD100_comparison} reports the median and IQR of the mean fitness of the comparison generations, together with the results from comparing the mean fitness of runs across obfuscation configurations.
The number of wins for each configuration is as follows: without obfuscation (13 wins), $O$ (12 wins), $T_{15}$ (11 wins), $A_{95}$ (10 wins), $Q_{20}$ (9 wins), $A_{90}$ (7 wins), $T_{30}$ (7 wins), $Q_{10}$ (6 wins), $A_{80}$ (5 wins), $B_{20}$ (3 wins), $T_{60}$ (3 wins), $B_{10}$ (2 wins), $Q_5$ (1 win), and $B_5$ (0 wins).
The runs with $T_{15}$ perform best among the configurable obfuscation methods, while those with \emph{fitness buckets} obfuscation perform among the worst.

\begin{table}
    \centering
    \caption{Median and interquartile range (IQR) of the mean fitness of the comparison generation across 31 runs for each obfuscation configuration on Instance \emph{kroD100} with an optimal fitness of \numprint{21294}. Matched-pairs rank-biserial correlations, rounded to one decimal place, are reported for comparing the performance of the configuration in each row with that in each column. Grey cells indicate statistically significant differences in performance. A dash (-) denotes no obfuscation.}
    \label{tab:tsp2_kroD100_comparison}
    \footnotesize
    \begin{tabular}{lrrrrrrrrrrrrrrr}
    \toprule
& \multirow{2}{*}{Median} & \multirow{2}{*}{IQR} & \multicolumn{13}{c}{Matched-Pairs Rank-Biserial Correlation} \\
\cmidrule(lr){4-16}
&        &     & - & $B_{20}$ & $B_{10}$ & $B_{5}$ & $A_{95}$ & $A_{90}$ & $A_{80}$ & $O$ & $Q_{20}$ & $Q_{10}$ & $Q_{5}$ & $T_{15}$ & $T_{30}$ \\
\midrule
- & \numprint{22579} & \numprint{948} \\
$B_{20}$ & \numprint{44944} & \numprint{3107} & \cellcolor[HTML]{D3D3D3} \numprint{1.0} \\
$B_{10}$ & \numprint{47218} & \numprint{1987} & \cellcolor[HTML]{D3D3D3} \numprint{1.0} & \cellcolor[HTML]{D3D3D3} \numprint{0.7} \\
$B_{5}$ & \numprint{54565} & \numprint{3585} & \cellcolor[HTML]{D3D3D3} \numprint{1.0} & \cellcolor[HTML]{D3D3D3} \numprint{1.0} & \cellcolor[HTML]{D3D3D3} \numprint{1.0} \\
$A_{95}$ & \numprint{31172} & \numprint{2207} & \cellcolor[HTML]{D3D3D3} \numprint{1.0} & \cellcolor[HTML]{D3D3D3} -\numprint{1.0} & \cellcolor[HTML]{D3D3D3} -\numprint{1.0} & \cellcolor[HTML]{D3D3D3} -\numprint{1.0} \\
$A_{90}$ & \numprint{33698} & \numprint{2368} & \cellcolor[HTML]{D3D3D3} \numprint{1.0} & \cellcolor[HTML]{D3D3D3} -\numprint{1.0} & \cellcolor[HTML]{D3D3D3} -\numprint{1.0} & \cellcolor[HTML]{D3D3D3} -\numprint{1.0} & \cellcolor[HTML]{D3D3D3} \numprint{1.0} \\
$A_{80}$ & \numprint{41366} & \numprint{3595} & \cellcolor[HTML]{D3D3D3} \numprint{1.0} & \cellcolor[HTML]{D3D3D3} -\numprint{0.8} & \cellcolor[HTML]{D3D3D3} -\numprint{1.0} & \cellcolor[HTML]{D3D3D3} -\numprint{1.0} & \cellcolor[HTML]{D3D3D3} \numprint{1.0} & \cellcolor[HTML]{D3D3D3} \numprint{1.0} \\
$O$ & \numprint{25795} & \numprint{875} & \cellcolor[HTML]{D3D3D3} \numprint{1.0} & \cellcolor[HTML]{D3D3D3} -\numprint{1.0} & \cellcolor[HTML]{D3D3D3} -\numprint{1.0} & \cellcolor[HTML]{D3D3D3} -\numprint{1.0} & \cellcolor[HTML]{D3D3D3} -\numprint{1.0} & \cellcolor[HTML]{D3D3D3} -\numprint{1.0} & \cellcolor[HTML]{D3D3D3} -\numprint{1.0} \\
$Q_{20}$ & \numprint{31864} & \numprint{1029} & \cellcolor[HTML]{D3D3D3} \numprint{1.0} & \cellcolor[HTML]{D3D3D3} -\numprint{1.0} & \cellcolor[HTML]{D3D3D3} -\numprint{1.0} & \cellcolor[HTML]{D3D3D3} -\numprint{1.0} & \cellcolor[HTML]{D3D3D3} \numprint{0.7} & \cellcolor[HTML]{D3D3D3} -\numprint{0.8} & \cellcolor[HTML]{D3D3D3} -\numprint{1.0} & \cellcolor[HTML]{D3D3D3} \numprint{1.0} \\
$Q_{10}$ & \numprint{37257} & \numprint{1828} & \cellcolor[HTML]{D3D3D3} \numprint{1.0} & \cellcolor[HTML]{D3D3D3} -\numprint{1.0} & \cellcolor[HTML]{D3D3D3} -\numprint{1.0} & \cellcolor[HTML]{D3D3D3} -\numprint{1.0} & \cellcolor[HTML]{D3D3D3} \numprint{1.0} & \cellcolor[HTML]{D3D3D3} \numprint{1.0} & \cellcolor[HTML]{D3D3D3} -\numprint{0.9} & \cellcolor[HTML]{D3D3D3} \numprint{1.0} & \cellcolor[HTML]{D3D3D3} \numprint{1.0} \\
$Q_{5}$ & \numprint{48088} & \numprint{2940} & \cellcolor[HTML]{D3D3D3} \numprint{1.0} & \cellcolor[HTML]{D3D3D3} \numprint{0.9} & \cellcolor[HTML]{D3D3D3} \numprint{0.5} & \cellcolor[HTML]{D3D3D3} -\numprint{1.0} & \cellcolor[HTML]{D3D3D3} \numprint{1.0} & \cellcolor[HTML]{D3D3D3} \numprint{1.0} & \cellcolor[HTML]{D3D3D3} \numprint{1.0} & \cellcolor[HTML]{D3D3D3} \numprint{1.0} & \cellcolor[HTML]{D3D3D3} \numprint{1.0} & \cellcolor[HTML]{D3D3D3} \numprint{1.0} \\
$T_{15}$ & \numprint{28621} & \numprint{1216} & \cellcolor[HTML]{D3D3D3} \numprint{1.0} & \cellcolor[HTML]{D3D3D3} -\numprint{1.0} & \cellcolor[HTML]{D3D3D3} -\numprint{1.0} & \cellcolor[HTML]{D3D3D3} -\numprint{1.0} & \cellcolor[HTML]{D3D3D3} -\numprint{0.8} & \cellcolor[HTML]{D3D3D3} -\numprint{1.0} & \cellcolor[HTML]{D3D3D3} -\numprint{1.0} & \cellcolor[HTML]{D3D3D3} \numprint{1.0} & \cellcolor[HTML]{D3D3D3} -\numprint{1.0} & \cellcolor[HTML]{D3D3D3} -\numprint{1.0} & \cellcolor[HTML]{D3D3D3} -\numprint{1.0} \\
$T_{30}$ & \numprint{33382} & \numprint{2705} & \cellcolor[HTML]{D3D3D3} \numprint{1.0} & \cellcolor[HTML]{D3D3D3} -\numprint{1.0} & \cellcolor[HTML]{D3D3D3} -\numprint{1.0} & \cellcolor[HTML]{D3D3D3} -\numprint{1.0} & \cellcolor[HTML]{D3D3D3} \numprint{0.9} & -\numprint{0.4} & \cellcolor[HTML]{D3D3D3} -\numprint{1.0} & \cellcolor[HTML]{D3D3D3} \numprint{1.0} & \cellcolor[HTML]{D3D3D3} \numprint{0.6} & \cellcolor[HTML]{D3D3D3} -\numprint{1.0} & \cellcolor[HTML]{D3D3D3} -\numprint{1.0} & \cellcolor[HTML]{D3D3D3} \numprint{1.0} \\
$T_{60}$ & \numprint{43925} & \numprint{3748} & \cellcolor[HTML]{D3D3D3} \numprint{1.0} & -\numprint{0.2} & \cellcolor[HTML]{D3D3D3} -\numprint{0.9} & \cellcolor[HTML]{D3D3D3} -\numprint{1.0} & \cellcolor[HTML]{D3D3D3} \numprint{1.0} & \cellcolor[HTML]{D3D3D3} \numprint{1.0} & \cellcolor[HTML]{D3D3D3} \numprint{0.5} & \cellcolor[HTML]{D3D3D3} \numprint{1.0} & \cellcolor[HTML]{D3D3D3} \numprint{1.0} & \cellcolor[HTML]{D3D3D3} \numprint{1.0} & \cellcolor[HTML]{D3D3D3} -\numprint{1.0} & \cellcolor[HTML]{D3D3D3} \numprint{1.0} & \cellcolor[HTML]{D3D3D3} \numprint{1.0} \\

    \bottomrule
    \end{tabular}
\end{table}

The median performance of Generation~500 differs across obfuscation configurations, with at least one configuration within each method achieving performance close to the optimum.
When imposing the time budget on evaluation, the results are similar to those on the AP instance.
Again, although reducing information leakage, the results show that obfuscation indeed negatively impacts performance.
The binary fitness information provided by $T_{15}$ enables the genetic algorithm to find solution sets with a median of fitness means of 134.41~\% of the optimum within the time budget.
While the performance may intuitively seem poor, be aware that the initial populations without obfuscation have a median performance of 663.04~\% of the optimum.

\subsection{Multi-Objective Assignment Problem (with One Objective Obfuscated)}
\begin{figure}
    \centering
    \includegraphics[width=\linewidth]{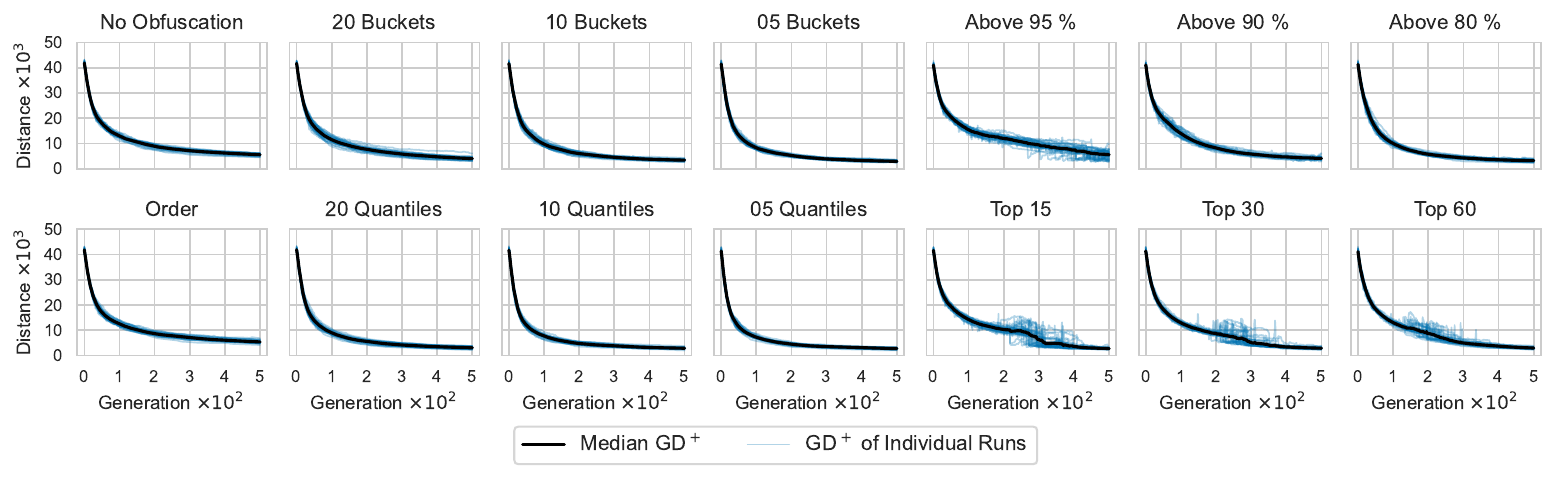}
    \caption{Median GD\textsuperscript{+} per generation and GD\textsuperscript{+} of 31 individual runs over 500 generations for each obfuscation configuration on Instance \emph{ap2K100x100} with one objective obfuscated.}
    \label{fig:moap2_ap2K100x100_convergence_gd_plus}
\end{figure}

We present the results of NSGA-II on Instance \emph{ap2K100x100} with the fitness values for one objective, namely, Objective~1, obfuscated.
Figure~\ref{fig:moap2_ap2K100x100_convergence_gd_plus} shows the median GD\textsuperscript{+} per generation and the GD\textsuperscript{+} of 31 individual runs over 500 generations for each obfuscation configuration.
Individual runs with $A_{95}$, $T_{15}$, $T_{30}$ and $T_{60}$ deviate considerably from the median curve and exhibit sharp increases.
For $T_{15}$, $T_{30}$, and $T_{60}$, this behavior eventually stabilizes.
Individual runs with the remaining obfuscation configurations closely follow the median curve, indicating low variability across runs.

\begin{figure}
    \centering
    \includegraphics[width=\linewidth]{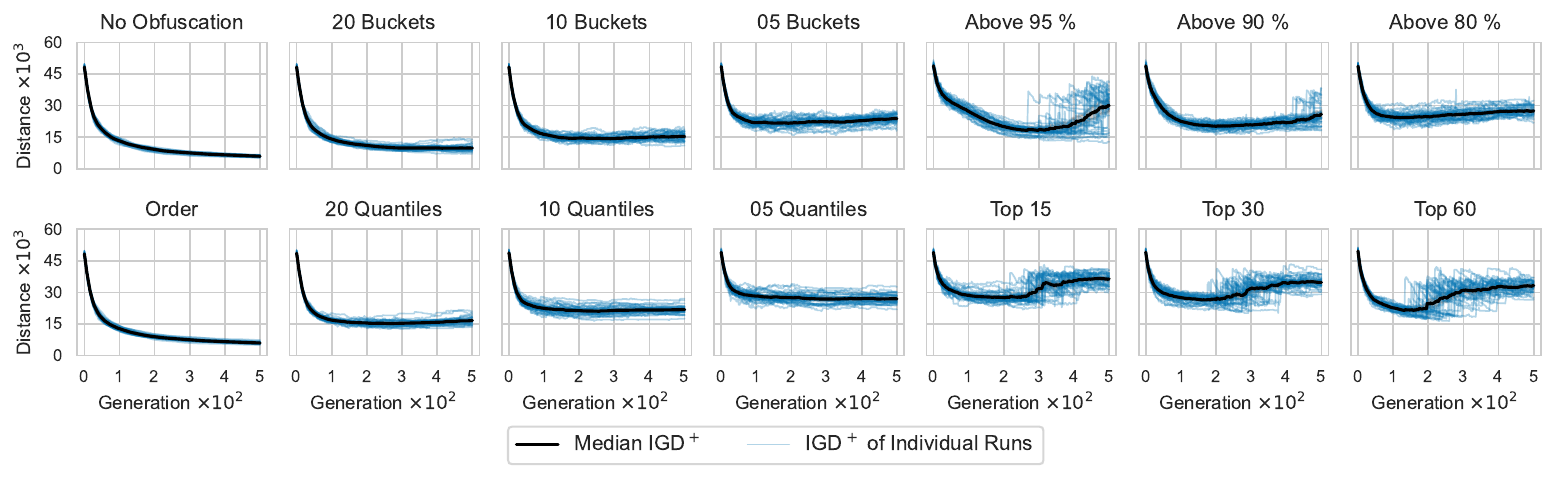}
    \caption{Median IGD\textsuperscript{+} per generation and IGD\textsuperscript{+} of 31 individual runs over 500 generations for each obfuscation configuration on Instance \emph{ap2K100x100} with one objective obfuscated.}
    \label{fig:moap2_ap2K100x100_convergence_igd_plus}
\end{figure}

Figure~\ref{fig:moap2_ap2K100x100_convergence_igd_plus} shows the median IGD\textsuperscript{+} per generation and the IGD\textsuperscript{+} of 31 individual runs over 500 generations for each obfuscation configuration.
Individual runs with \emph{fitness buckets} and \emph{order quantiles} obfuscation stagnate early and deviate considerably from the median curve.
The curve of individual runs with \emph{above threshold} and \emph{top individuals} obfuscation even increases at some point.
These runs deviate substantially from the median curve and exhibit sharp increases.

Table~\ref{tab:moap2_summary} reports the median and IQR of GD\textsuperscript{+} and IGD\textsuperscript{+} across 31 runs for the initial population and Generation~500 for each obfuscation configuration.
Within each obfuscation method, the median GD\textsuperscript{+} of Generation~500 tends to improve, i.e., decrease, when the number of buckets, quantiles, or top individuals is reduced, or the threshold is lowered.
The median IGD\textsuperscript{+} of Generation~500 tends to improve, i.e., decrease, when the number of buckets, quantiles, or top individuals is increased.
The results for \emph{above threshold} obfuscation indicate no trend based on the median IGD\textsuperscript{+}.

\begin{table}
    \caption{Median and interquartile range (IQR) of GD\textsuperscript{+} and IGD\textsuperscript{+}, rounded to integers, for the initial population and Generation 500 across 31 runs for each obfuscation configuration on Instance \emph{ap2K100x100} with one objective obfuscated.}
    \footnotesize
    \label{tab:moap2_summary}
    \begin{tabular}{rrrrrrrrr}
        \toprule
         \multicolumn{1}{c}{\multirow{3}{*}{Obfuscation}} & \multicolumn{4}{c}{GD\textsuperscript{+}} & \multicolumn{4}{c}{IGD\textsuperscript{+}} \\
         \cmidrule(lr){2-5}
         \cmidrule(lr){6-9}
         & \multicolumn{2}{c}{Initial Population} & \multicolumn{2}{c}{Generation 500} & \multicolumn{2}{c}{Initial Population} &  \multicolumn{2}{c}{Generation 500} \\
         \cmidrule(lr){2-3}
         \cmidrule(lr){4-5}
         \cmidrule(lr){6-7}
         \cmidrule(lr){8-9}
         & \multicolumn{1}{c}{Median} & \multicolumn{1}{c}{IQR}& \multicolumn{1}{c}{Median} & \multicolumn{1}{c}{IQR}& \multicolumn{1}{c}{Median} & \multicolumn{1}{c}{IQR}& \multicolumn{1}{c}{Median} & \multicolumn{1}{c}{IQR} \\
         \midrule

No Obfuscation & \numprint{42018} & \numprint{1333} & \numprint{5597} & \numprint{546} & \numprint{48319} & \numprint{1552} & \numprint{5951} & \numprint{566} \\
20 Buckets & \numprint{42001} & \numprint{1173} & \numprint{3949} & \numprint{745}& \numprint{48319} & \numprint{1552} & \numprint{9693} & \numprint{1554} \\
10 Buckets & \numprint{41679} & \numprint{1161} & \numprint{3467} & \numprint{411} & \numprint{48410} & \numprint{1737} & \numprint{15206} & \numprint{3070} \\
5 Buckets & \numprint{41319} & \numprint{1180} & \numprint{2876} & \numprint{448} & \numprint{48410} & \numprint{1959} & \numprint{23900} & \numprint{3179} \\
Above 95 \% & \numprint{41065} & \numprint{1102} & \numprint{5938} & \numprint{2667} & \numprint{48923} & \numprint{2105} & \numprint{33066} & \numprint{8633} \\
Above 90 \% & \numprint{41081} & \numprint{1204} & \numprint{4004} & \numprint{684} & \numprint{48876} & \numprint{2095} & \numprint{24408} & \numprint{7341} \\
Above 80 \% & \numprint{41280} & \numprint{1297} & \numprint{3218} & \numprint{624} & \numprint{48410} & \numprint{2220} & \numprint{27225} & \numprint{3415} \\
Order & \numprint{42018} & \numprint{1333} & \numprint{5428} & \numprint{629} & \numprint{48319} & \numprint{1552} & \numprint{5901} & \numprint{485} \\
20 Quantiles & \numprint{42308} & \numprint{1490} & \numprint{3029} & \numprint{744} & \numprint{48806} & \numprint{1767} & \numprint{16573} & \numprint{2525} \\
10 Quantiles & \numprint{41834} & \numprint{1174} & \numprint{2887} & \numprint{383} & \numprint{48836} & \numprint{1438} & \numprint{21858} & \numprint{2154} \\
5 Quantiles & \numprint{41460} & \numprint{1319} & \numprint{2783} & \numprint{420} & \numprint{49074} & \numprint{1494} & \numprint{26248} & \numprint{3364} \\
Top 15 & \numprint{41753} & \numprint{1343} & \numprint{2756} & \numprint{395} & \numprint{49190} & \numprint{1777} & \numprint{36590} & \numprint{3865} \\
Top 30 & \numprint{41511} & \numprint{1114} & \numprint{2849} & \numprint{428} & \numprint{49221} & \numprint{1751} & \numprint{34609} & \numprint{3441} \\
Top 60 & \numprint{41152} & \numprint{1548} & \numprint{3015} & \numprint{475} & \numprint{49652} & \numprint{1383} & \numprint{33774} & \numprint{3184} \\

         \bottomrule
    \end{tabular}
\end{table}

In contrast to the results for the AP and TSP instances in Table~\ref{tab:ap1K100x100_kroD100_summary}, the performance of initial populations in Table~\ref{tab:moap2_summary} differs between corresponding configurations of \emph{order quantiles} and \emph{top individuals} obfuscation.
The performance on the MOAP is measured using the set of estimated non-dominated individuals instead of the set of estimated optimal individuals.
With \emph{top individuals} obfuscation providing only two distinct fitness values for obfuscated objectives, \emph{order quantiles} obfuscation provides more distinct values, resulting in different sets of estimated non-dominated solutions.

\begin{figure}
    \centering
    \includegraphics[width=\linewidth]{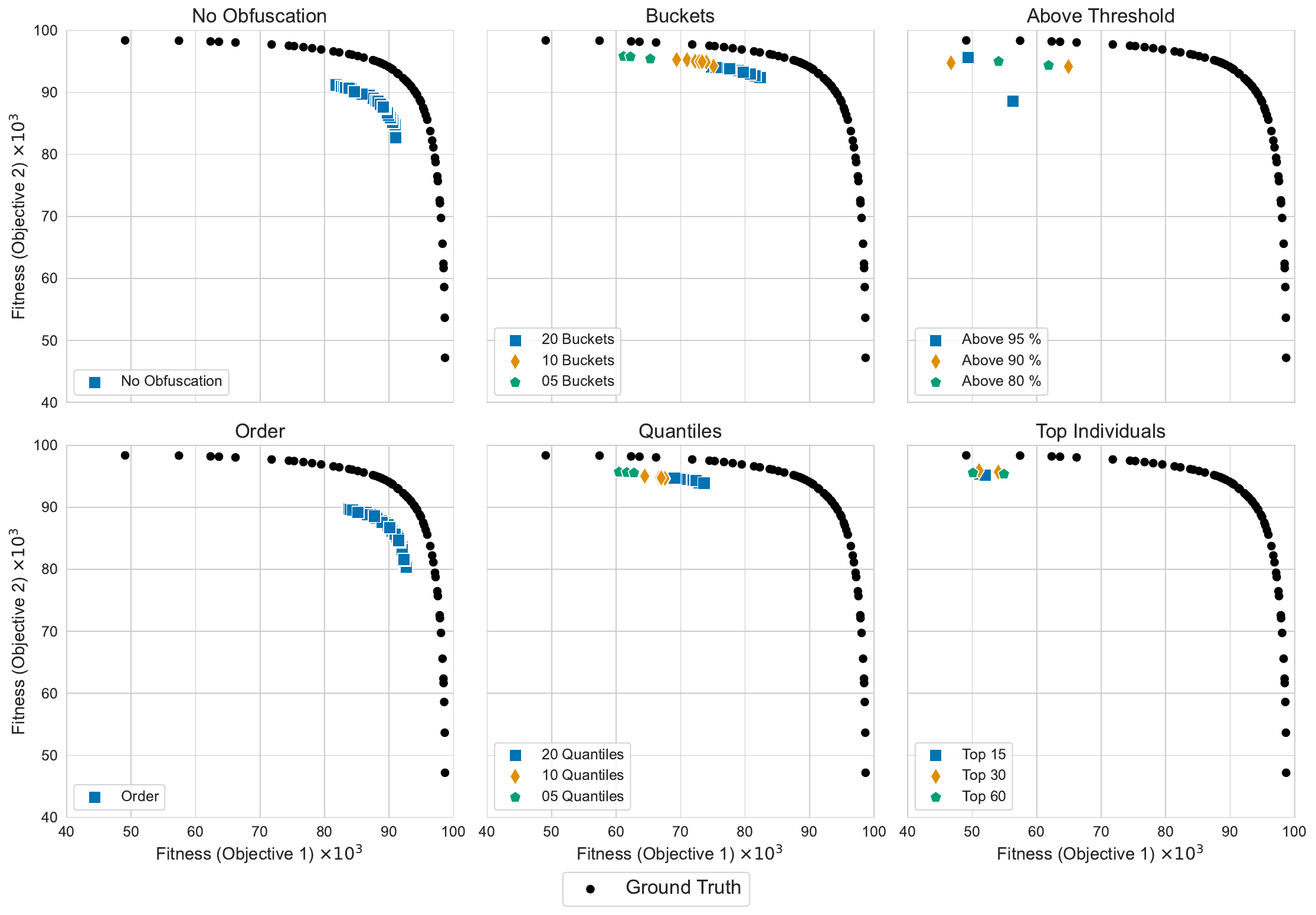}
    \caption{Set of estimated non-dominated solution sets of Generation 500 from the run with the median IGD\textsuperscript{+} for each obfuscation configuration on Instance \emph{ap2K100x100} with one objective (Objective~1) obfuscated.}
    \label{fig:moap2_ap2K100x100_results}
\end{figure}

The results in Table~\ref{tab:moap2_summary} show that, after 500 generations, runs with obfuscation achieve a lower median GD\textsuperscript{+} than runs without obfuscation, except for $A_{95}$, but perform poorly in terms of median IGD\textsuperscript{+}.
Figure~\ref{fig:moap2_ap2K100x100_results} shows the set of estimated non-dominated solutions of Generation~500 from the run with the median IGD\textsuperscript{+} for each obfuscation configuration.
While the solution sets obtained by runs without obfuscation and with \emph{order} obfuscation are balanced between the two objectives, those obtained with the remaining obfuscation configurations are skewed towards the non-obfuscated objective, i.e., Objective~2.
Since subsets of solutions have the same estimated fitness for the obfuscated objective, they only vary in fitness for the non-obfuscated objective.
Therefore, non-dominated sorting ranks solutions within the same subset higher if they perform better in the non-obfuscated objective.

Table~\ref{tab:moap2_ap2K100x100_comparison_gd_plus} reports the median and IQR of the GD\textsuperscript{+} of the comparison generations, together with the results from comparing the GD\textsuperscript{+} of runs across obfuscation configurations.
The number of wins for each configuration is as follows: without obfuscation (13 wins), $A_{80}$ (12 wins), $Q_5$ (11 wins), $Q_{10}$ (10 wins), $A_{90}$ (8 wins), $Q_{20}$ (8 wins), $O$ (5 wins), $T_{30}$ (5 wins), $T_{60}$ (5 wins), $T_{15}$ (4 wins), $B_5$ (2 wins), $A_{95}$ (2 wins), $B_{10}$ (1 win), and $B_{20}$ (0 wins).
The runs with $A_{80}$, $Q_5$ and $Q_{10}$ perform among the best and even outperform those with \emph{order obfuscation}.
The runs with \emph{fitness buckets} obfuscation perform among the worst.

\begin{table}
    \centering
    \caption{Median and interquartile range (IQR) of GD\textsuperscript{+} of the comparison generation across 31 runs for each obfuscation configuration on Instance \emph{ap2K100x100} with one objective obfuscated. Matched-pairs rank-biserial correlations, rounded to one decimal place, are reported for comparing the performance of the configuration in each row with that in each column. Grey cells indicate statistically significant differences in performance. A dash (-) denotes no obfuscation.}
    \label{tab:moap2_ap2K100x100_comparison_gd_plus}
    \footnotesize
    \begin{tabular}{lrrrrrrrrrrrrrrr}
    \toprule
& \multirow{2}{*}{Median} & \multirow{2}{*}{IQR} & \multicolumn{13}{c}{Matched-Pairs Rank-Biserial Correlation} \\
\cmidrule(lr){4-16}
&        &     & - & $B_{20}$ & $B_{10}$ & $B_{5}$ & $A_{95}$ & $A_{90}$ & $A_{80}$ & $O$ & $Q_{20}$ & $Q_{10}$ & $Q_{5}$ & $T_{15}$ & $T_{30}$ \\
\midrule
- & \numprint{5597} & \numprint{546} \\
$B_{20}$ & \numprint{15962} & \numprint{2039} & \cellcolor[HTML]{D3D3D3} \numprint{1.0} \\
$B_{10}$ & \numprint{14403} & \numprint{997} & \cellcolor[HTML]{D3D3D3} \numprint{1.0} & \cellcolor[HTML]{D3D3D3} -\numprint{0.9} \\
$B_{5}$ & \numprint{12527} & \numprint{967} & \cellcolor[HTML]{D3D3D3} \numprint{1.0} & \cellcolor[HTML]{D3D3D3} -\numprint{1.0} & \cellcolor[HTML]{D3D3D3} -\numprint{0.9} \\
$A_{95}$ & \numprint{12523} & \numprint{1277} & \cellcolor[HTML]{D3D3D3} \numprint{1.0} & \cellcolor[HTML]{D3D3D3} -\numprint{1.0} & \cellcolor[HTML]{D3D3D3} -\numprint{0.9} & -\numprint{0.1} \\
$A_{90}$ & \numprint{9020} & \numprint{915} & \cellcolor[HTML]{D3D3D3} \numprint{1.0} & \cellcolor[HTML]{D3D3D3} -\numprint{1.0} & \cellcolor[HTML]{D3D3D3} -\numprint{1.0} & \cellcolor[HTML]{D3D3D3} -\numprint{1.0} & \cellcolor[HTML]{D3D3D3} -\numprint{1.0} \\
$A_{80}$ & \numprint{6198} & \numprint{677} & \cellcolor[HTML]{D3D3D3} \numprint{0.9} & \cellcolor[HTML]{D3D3D3} -\numprint{1.0} & \cellcolor[HTML]{D3D3D3} -\numprint{1.0} & \cellcolor[HTML]{D3D3D3} -\numprint{1.0} & \cellcolor[HTML]{D3D3D3} -\numprint{1.0} & \cellcolor[HTML]{D3D3D3} -\numprint{1.0} \\
$O$ & \numprint{9469} & \numprint{801} & \cellcolor[HTML]{D3D3D3} \numprint{1.0} & \cellcolor[HTML]{D3D3D3} -\numprint{1.0} & \cellcolor[HTML]{D3D3D3} -\numprint{1.0} & \cellcolor[HTML]{D3D3D3} -\numprint{1.0} & \cellcolor[HTML]{D3D3D3} -\numprint{1.0} & \cellcolor[HTML]{D3D3D3} \numprint{0.6} & \cellcolor[HTML]{D3D3D3} \numprint{1.0} \\
$Q_{20}$ & \numprint{8536} & \numprint{896} & \cellcolor[HTML]{D3D3D3} \numprint{1.0} & \cellcolor[HTML]{D3D3D3} -\numprint{1.0} & \cellcolor[HTML]{D3D3D3} -\numprint{1.0} & \cellcolor[HTML]{D3D3D3} -\numprint{1.0} & \cellcolor[HTML]{D3D3D3} -\numprint{1.0} & -\numprint{0.4} & \cellcolor[HTML]{D3D3D3} \numprint{1.0} & \cellcolor[HTML]{D3D3D3} -\numprint{0.8} \\
$Q_{10}$ & \numprint{7311} & \numprint{627} & \cellcolor[HTML]{D3D3D3} \numprint{1.0} & \cellcolor[HTML]{D3D3D3} -\numprint{1.0} & \cellcolor[HTML]{D3D3D3} -\numprint{1.0} & \cellcolor[HTML]{D3D3D3} -\numprint{1.0} & \cellcolor[HTML]{D3D3D3} -\numprint{1.0} & \cellcolor[HTML]{D3D3D3} -\numprint{1.0} & \cellcolor[HTML]{D3D3D3} \numprint{1.0} & \cellcolor[HTML]{D3D3D3} -\numprint{1.0} & \cellcolor[HTML]{D3D3D3} -\numprint{1.0} \\
$Q_{5}$ & \numprint{6818} & \numprint{539} & \cellcolor[HTML]{D3D3D3} \numprint{1.0} & \cellcolor[HTML]{D3D3D3} -\numprint{1.0} & \cellcolor[HTML]{D3D3D3} -\numprint{1.0} & \cellcolor[HTML]{D3D3D3} -\numprint{1.0} & \cellcolor[HTML]{D3D3D3} -\numprint{1.0} & \cellcolor[HTML]{D3D3D3} -\numprint{1.0} & \cellcolor[HTML]{D3D3D3} \numprint{0.8} & \cellcolor[HTML]{D3D3D3} -\numprint{1.0} & \cellcolor[HTML]{D3D3D3} -\numprint{1.0} & \cellcolor[HTML]{D3D3D3} -\numprint{0.5} \\
$T_{15}$ & \numprint{11059} & \numprint{1513} & \cellcolor[HTML]{D3D3D3} \numprint{1.0} & \cellcolor[HTML]{D3D3D3} -\numprint{1.0} & \cellcolor[HTML]{D3D3D3} -\numprint{1.0} & \cellcolor[HTML]{D3D3D3} -\numprint{1.0} & \cellcolor[HTML]{D3D3D3} -\numprint{0.9} & \cellcolor[HTML]{D3D3D3} \numprint{1.0} & \cellcolor[HTML]{D3D3D3} \numprint{1.0} & \cellcolor[HTML]{D3D3D3} \numprint{0.9} & \cellcolor[HTML]{D3D3D3} \numprint{1.0} & \cellcolor[HTML]{D3D3D3} \numprint{1.0} & \cellcolor[HTML]{D3D3D3} \numprint{1.0} \\
$T_{30}$ & \numprint{9547} & \numprint{795} & \cellcolor[HTML]{D3D3D3} \numprint{1.0} & \cellcolor[HTML]{D3D3D3} -\numprint{1.0} & \cellcolor[HTML]{D3D3D3} -\numprint{1.0} & \cellcolor[HTML]{D3D3D3} -\numprint{1.0} & \cellcolor[HTML]{D3D3D3} -\numprint{1.0} & \cellcolor[HTML]{D3D3D3} \numprint{0.6} & \cellcolor[HTML]{D3D3D3} \numprint{1.0} & \numprint{0.2} & \cellcolor[HTML]{D3D3D3} \numprint{1.0} & \cellcolor[HTML]{D3D3D3} \numprint{1.0} & \cellcolor[HTML]{D3D3D3} \numprint{1.0} & \cellcolor[HTML]{D3D3D3} -\numprint{0.9} \\
$T_{60}$ & \numprint{9843} & \numprint{1439} & \cellcolor[HTML]{D3D3D3} \numprint{1.0} & \cellcolor[HTML]{D3D3D3} -\numprint{1.0} & \cellcolor[HTML]{D3D3D3} -\numprint{1.0} & \cellcolor[HTML]{D3D3D3} -\numprint{1.0} & \cellcolor[HTML]{D3D3D3} -\numprint{0.9} & \cellcolor[HTML]{D3D3D3} \numprint{0.6} & \cellcolor[HTML]{D3D3D3} \numprint{1.0} & \numprint{0.4} & \cellcolor[HTML]{D3D3D3} \numprint{0.9} & \cellcolor[HTML]{D3D3D3} \numprint{1.0} & \cellcolor[HTML]{D3D3D3} \numprint{1.0} & \cellcolor[HTML]{D3D3D3} -\numprint{0.7} & \numprint{0.3} \\

    \bottomrule
    \end{tabular}
\end{table}

Table~\ref{tab:moap2_ap2K100x100_comparison_igd_plus} reports the median and IQR of the IGD\textsuperscript{+} of the comparison generation, together with the results from comparing the IGD\textsuperscript{+} of runs across obfuscation configurations.
The number of wins for each configuration is as follows: without obfuscation (13 wins), $O$ (12 wins), $Q_{20}$ (11 wins), $B_{20}$ (10 wins), $B_{10}$ (8 wins), $A_{95}$ (6 wins), $A_{90}$ (6 wins), $Q_{10}$ (5 wins), $T_{60}$ (5 wins), $B_5$ (3 wins), $A_{80}$ (3 wins), $T_{30}$ (1 win), $Q_5$ (0 wins), and $T_{15}$ (0 wins).
The runs with $Q_{20}$ and $B_{20}$ perform among the best due to providing the most information about fitness among the configurable obfuscation methods.
The runs with \emph{top individuals} obfuscation perform among the worst.

\begin{table}
    \centering
    \caption{Median and interquartile range (IQR) of IGD\textsuperscript{+} of the comparison generation across 31 runs for each obfuscation configuration on Instance \emph{ap2K100x100} with one objective obfuscated. Matched-pairs rank-biserial correlations, rounded to one decimal place, are reported for comparing the performance of the configuration in each row with that in each column. Grey cells indicate statistically significant differences in performance. A dash (-) denotes no obfuscation.}
    \label{tab:moap2_ap2K100x100_comparison_igd_plus}
    \footnotesize
    \begin{tabular}{lrrrrrrrrrrrrrrr}
    \toprule
& \multirow{2}{*}{Median} & \multirow{2}{*}{IQR} & \multicolumn{13}{c}{Matched-Pairs Rank-Biserial Correlation} \\
\cmidrule(lr){4-16}
&        &     & - & $B_{20}$ & $B_{10}$ & $B_{5}$ & $A_{95}$ & $A_{90}$ & $A_{80}$ & $O$ & $Q_{20}$ & $Q_{10}$ & $Q_{5}$ & $T_{15}$ & $T_{30}$ \\
\midrule
- & \numprint{5951} & \numprint{566} \\
$B_{20}$ & \numprint{18072} & \numprint{1179} & \cellcolor[HTML]{D3D3D3} \numprint{1.0} \\
$B_{10}$ & \numprint{19733} & \numprint{1559} & \cellcolor[HTML]{D3D3D3} \numprint{1.0} & \cellcolor[HTML]{D3D3D3} \numprint{0.9} \\
$B_{5}$ & \numprint{24627} & \numprint{1983} & \cellcolor[HTML]{D3D3D3} \numprint{1.0} & \cellcolor[HTML]{D3D3D3} \numprint{1.0} & \cellcolor[HTML]{D3D3D3} \numprint{1.0} \\
$A_{95}$ & \numprint{21236} & \numprint{2976} & \cellcolor[HTML]{D3D3D3} \numprint{1.0} & \cellcolor[HTML]{D3D3D3} \numprint{0.9} & \cellcolor[HTML]{D3D3D3} \numprint{0.4} & \cellcolor[HTML]{D3D3D3} -\numprint{0.9} \\
$A_{90}$ & \numprint{20586} & \numprint{2307} & \cellcolor[HTML]{D3D3D3} \numprint{1.0} & \cellcolor[HTML]{D3D3D3} \numprint{0.9} & \numprint{0.2} & \cellcolor[HTML]{D3D3D3} -\numprint{1.0} & -\numprint{0.2} \\
$A_{80}$ & \numprint{24844} & \numprint{3283} & \cellcolor[HTML]{D3D3D3} \numprint{1.0} & \cellcolor[HTML]{D3D3D3} \numprint{1.0} & \cellcolor[HTML]{D3D3D3} \numprint{1.0} & \numprint{0.1} & \cellcolor[HTML]{D3D3D3} \numprint{0.9} & \cellcolor[HTML]{D3D3D3} \numprint{1.0} \\
$O$ & \numprint{9663} & \numprint{805} & \cellcolor[HTML]{D3D3D3} \numprint{1.0} & \cellcolor[HTML]{D3D3D3} -\numprint{1.0} & \cellcolor[HTML]{D3D3D3} -\numprint{1.0} & \cellcolor[HTML]{D3D3D3} -\numprint{1.0} & \cellcolor[HTML]{D3D3D3} -\numprint{1.0} & \cellcolor[HTML]{D3D3D3} -\numprint{1.0} & \cellcolor[HTML]{D3D3D3} -\numprint{1.0} \\
$Q_{20}$ & \numprint{16450} & \numprint{1200} & \cellcolor[HTML]{D3D3D3} \numprint{1.0} & \cellcolor[HTML]{D3D3D3} -\numprint{0.8} & \cellcolor[HTML]{D3D3D3} -\numprint{1.0} & \cellcolor[HTML]{D3D3D3} -\numprint{1.0} & \cellcolor[HTML]{D3D3D3} -\numprint{1.0} & \cellcolor[HTML]{D3D3D3} -\numprint{1.0} & \cellcolor[HTML]{D3D3D3} -\numprint{1.0} & \cellcolor[HTML]{D3D3D3} \numprint{1.0} \\
$Q_{10}$ & \numprint{22437} & \numprint{2137} & \cellcolor[HTML]{D3D3D3} \numprint{1.0} & \cellcolor[HTML]{D3D3D3} \numprint{1.0} & \cellcolor[HTML]{D3D3D3} \numprint{0.9} & \cellcolor[HTML]{D3D3D3} -\numprint{0.8} & \cellcolor[HTML]{D3D3D3} \numprint{0.4} & \cellcolor[HTML]{D3D3D3} \numprint{0.8} & \cellcolor[HTML]{D3D3D3} -\numprint{0.8} & \cellcolor[HTML]{D3D3D3} \numprint{1.0} & \cellcolor[HTML]{D3D3D3} \numprint{1.0} \\
$Q_{5}$ & \numprint{28034} & \numprint{2536} & \cellcolor[HTML]{D3D3D3} \numprint{1.0} & \cellcolor[HTML]{D3D3D3} \numprint{1.0} & \cellcolor[HTML]{D3D3D3} \numprint{1.0} & \cellcolor[HTML]{D3D3D3} \numprint{1.0} & \cellcolor[HTML]{D3D3D3} \numprint{1.0} & \cellcolor[HTML]{D3D3D3} \numprint{1.0} & \cellcolor[HTML]{D3D3D3} \numprint{0.9} & \cellcolor[HTML]{D3D3D3} \numprint{1.0} & \cellcolor[HTML]{D3D3D3} \numprint{1.0} & \cellcolor[HTML]{D3D3D3} \numprint{1.0} \\
$T_{15}$ & \numprint{27535} & \numprint{2394} & \cellcolor[HTML]{D3D3D3} \numprint{1.0} & \cellcolor[HTML]{D3D3D3} \numprint{1.0} & \cellcolor[HTML]{D3D3D3} \numprint{1.0} & \cellcolor[HTML]{D3D3D3} \numprint{0.9} & \cellcolor[HTML]{D3D3D3} \numprint{1.0} & \cellcolor[HTML]{D3D3D3} \numprint{1.0} & \cellcolor[HTML]{D3D3D3} \numprint{0.8} & \cellcolor[HTML]{D3D3D3} \numprint{1.0} & \cellcolor[HTML]{D3D3D3} \numprint{1.0} & \cellcolor[HTML]{D3D3D3} \numprint{1.0} & -\numprint{0.3} \\
$T_{30}$ & \numprint{26617} & \numprint{2460} & \cellcolor[HTML]{D3D3D3} \numprint{1.0} & \cellcolor[HTML]{D3D3D3} \numprint{1.0} & \cellcolor[HTML]{D3D3D3} \numprint{1.0} & \cellcolor[HTML]{D3D3D3} \numprint{0.9} & \cellcolor[HTML]{D3D3D3} \numprint{1.0} & \cellcolor[HTML]{D3D3D3} \numprint{1.0} & \cellcolor[HTML]{D3D3D3} \numprint{0.6} & \cellcolor[HTML]{D3D3D3} \numprint{1.0} & \cellcolor[HTML]{D3D3D3} \numprint{1.0} & \cellcolor[HTML]{D3D3D3} \numprint{0.9} & \cellcolor[HTML]{D3D3D3} -\numprint{0.7} & -\numprint{0.4} \\
$T_{60}$ & \numprint{21301} & \numprint{3500} & \cellcolor[HTML]{D3D3D3} \numprint{1.0} & \cellcolor[HTML]{D3D3D3} \numprint{0.9} & \cellcolor[HTML]{D3D3D3} \numprint{0.6} & \cellcolor[HTML]{D3D3D3} -\numprint{0.6} & \numprint{0.0} & \numprint{0.3} & \cellcolor[HTML]{D3D3D3} -\numprint{0.6} & \cellcolor[HTML]{D3D3D3} \numprint{1.0} & \cellcolor[HTML]{D3D3D3} \numprint{1.0} & -\numprint{0.3} & \cellcolor[HTML]{D3D3D3} -\numprint{0.9} & \cellcolor[HTML]{D3D3D3} -\numprint{0.8} & \cellcolor[HTML]{D3D3D3} -\numprint{0.7} \\

    \bottomrule
    \end{tabular}
\end{table}

Obfuscating one objective results in solution sets close to the Pareto front but skewed towards the non-obfuscated objective.
Although reducing information leakage, the results show that obfuscation indeed negatively impacts GD\textsuperscript{+} and IGD\textsuperscript{+} when imposing the time budged on evaluation.
Obfuscation configurations that perform well on GD\textsuperscript{+} may perform poorly on IGD\textsuperscript{+}, and vice versa.
When summing the comparison wins of GD\textsuperscript{+} and IGD\textsuperscript{+}, runs with $Q_{20}$ perform best among the configurable obfuscation methods within the time budget.

\subsection{Multi-Objective Assignment Problem (with Two Objectives Obfuscated)}
\begin{figure}
    \centering
    \includegraphics[width=\linewidth]{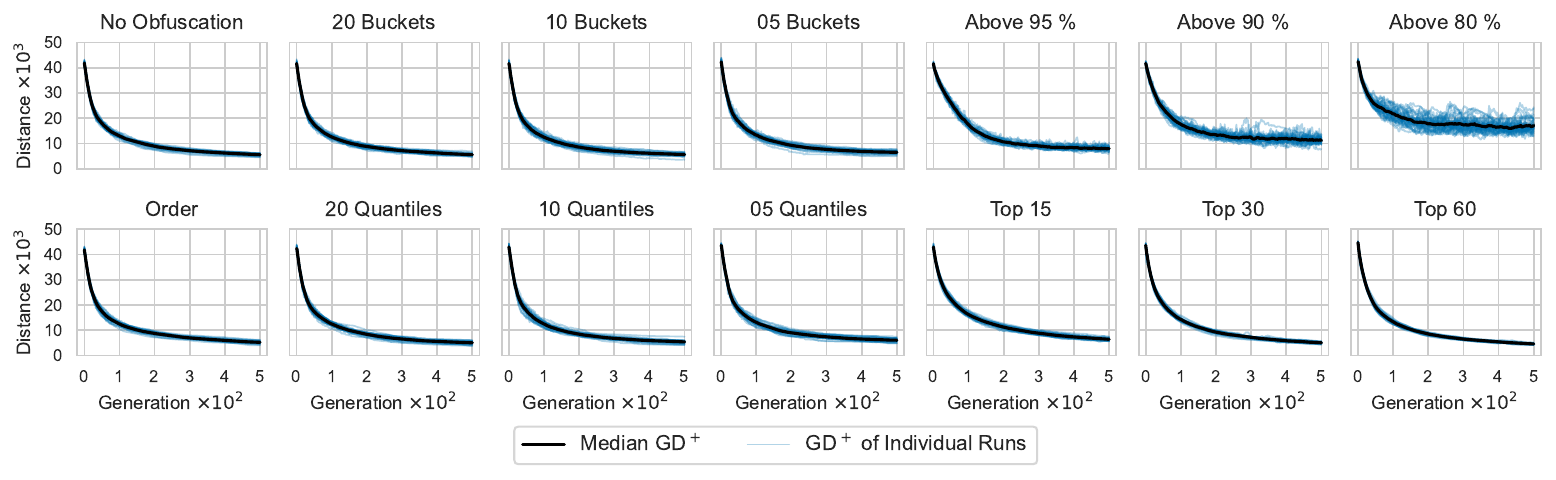}
    \caption{Median GD\textsuperscript{+} per generation and GD\textsuperscript{+} of 31 individual runs over 500 generations for each obfuscation configuration on Instance \emph{ap2K100x100} with two objectives obfuscated.}
    \label{fig:moap3_ap2K100x100_convergence_gd_plus}
\end{figure}

We present the results of NSGA-II on Instance \emph{ap2K100x100} with the fitness values for two objectives obfuscated.
Figure~\ref{fig:moap3_ap2K100x100_convergence_gd_plus} shows the median GD\textsuperscript{+} per generation and the GD\textsuperscript{+} of 31 individual runs over 500 generations for each obfuscation configuration.
Individual runs with \emph{above threshold} obfuscation deviate from the median curve, with deviations increasing from moderate with $A_{80}$ to considerable with $A_{90}$ and substantial with $A_{95}$.
Individual runs with the remaining obfuscation configurations closely follow the median curve, indicating low variability across runs.

\begin{figure}
    \centering
    \includegraphics[width=\linewidth]{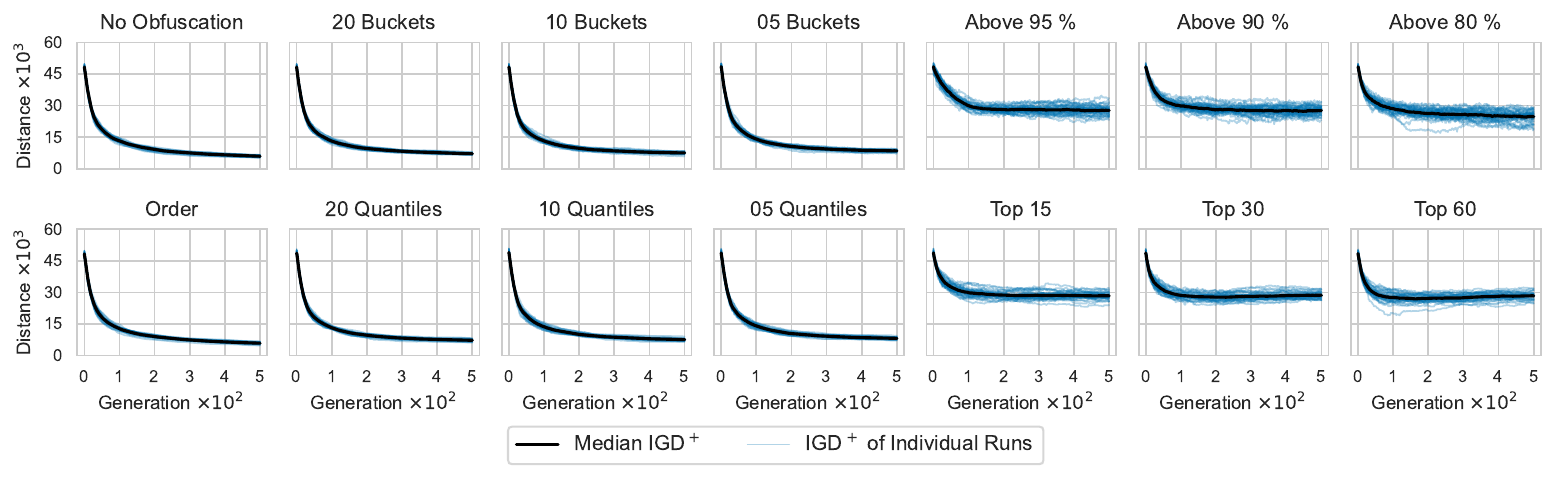}
    \caption{Median IGD\textsuperscript{+} per generation and IGD\textsuperscript{+} of 31 individual runs over 500 generations for each obfuscation configuration on Instance \emph{ap2K100x100} with two objectives obfuscated.}
    \label{fig:moap3_ap2K100x100_convergence_igd_plus}
\end{figure}

Figure~\ref{fig:moap3_ap2K100x100_convergence_igd_plus} shows the median IGD\textsuperscript{+} per generation and the individual IGD\textsuperscript{+} of 31 runs over 500 generations for each obfuscation configuration.
Individual runs with \emph{above threshold} and \emph{top individuals} obfuscation stagnate early and deviate considerably from the median curve.
Individual runs with the remaining obfuscation configurations closely follow the median curve, indicating low variability across runs.

Table~\ref{tab:moap3_summary} reports the median and IQR of GD\textsuperscript{+} and IGD\textsuperscript{+} across 31 runs for the initial population and Generation~500 for each obfuscation configuration.
Within each obfuscation method, the median GD\textsuperscript{+} of Generation~500 tends to improve, i.e., decrease, when the number of buckets, quantiles, or top individuals is increased, or the threshold is raised.
The median IGD\textsuperscript{+} of Generation~500 tends to improve, i.e., decrease, when the number of buckets or quantiles is increased and, to a small extent, when the number of top individuals is reduced.
The results for \emph{above threshold} obfuscation indicate no trend based on the median IGD\textsuperscript{+}.

\begin{table}
    \caption{Median and interquartile range (IQR) of GD\textsuperscript{+} and IGD\textsuperscript{+}, rounded to integers, for the initial population and Generation 500 across 31 runs for each obfuscation configuration on Instance \emph{ap2K100x100} with two objectives obfuscated.}
    \footnotesize
    \label{tab:moap3_summary}
    \begin{tabular}{rrrrrrrrr}
        \toprule
         \multicolumn{1}{c}{\multirow{3}{*}{Obfuscation}} & \multicolumn{4}{c}{GD\textsuperscript{+}} & \multicolumn{4}{c}{IGD\textsuperscript{+}} \\
         \cmidrule(lr){2-5}
         \cmidrule(lr){6-9}
         & \multicolumn{2}{c}{Initial Population} & \multicolumn{2}{c}{Generation 500} & \multicolumn{2}{c}{Initial Population} &  \multicolumn{2}{c}{Generation 500} \\
         \cmidrule(lr){2-3}
         \cmidrule(lr){4-5}
         \cmidrule(lr){6-7}
         \cmidrule(lr){8-9}
         & \multicolumn{1}{c}{Median} & \multicolumn{1}{c}{IQR}& \multicolumn{1}{c}{Median} & \multicolumn{1}{c}{IQR}& \multicolumn{1}{c}{Median} & \multicolumn{1}{c}{IQR}& \multicolumn{1}{c}{Median} & \multicolumn{1}{c}{IQR} \\
         \midrule

No Obfuscation & \numprint{42018} & \numprint{1333} & \numprint{5597} & \numprint{546} & \numprint{48319} & \numprint{1552} & \numprint{5951} & \numprint{566} \\
20 Buckets & \numprint{41819} & \numprint{1348} & \numprint{5627} & \numprint{557} & \numprint{48319} & \numprint{1561} & \numprint{7083} & \numprint{416} \\
10 Buckets & \numprint{41853} & \numprint{1731} & \numprint{5664} & \numprint{793} & \numprint{48410} & \numprint{1610} & \numprint{7529} & \numprint{623} \\
5 Buckets & \numprint{42405} & \numprint{1357} & \numprint{6602} & \numprint{868} & \numprint{48442} & \numprint{1667} & \numprint{8524} & \numprint{542} \\
Above 95 \% & \numprint{41574} & \numprint{489} & \numprint{7831} & \numprint{1162} & \numprint{48442} & \numprint{1494} & \numprint{27803} & \numprint{3295} \\
Above 90 \% & \numprint{41738} & \numprint{859} & \numprint{11242} & \numprint{1469} & \numprint{48319} & \numprint{1570} & \numprint{28082} & \numprint{2862} \\
Above 80 \% & \numprint{42572} & \numprint{921} & \numprint{16270} & \numprint{4044} & \numprint{48442} & \numprint{1392} & \numprint{25207} & \numprint{3669} \\
Order & \numprint{42018} & \numprint{1333} & \numprint{5224} & \numprint{550} & \numprint{48319} & \numprint{1552} & \numprint{5898} & \numprint{488} \\
20 Quantiles & \numprint{42597} & \numprint{1452} & \numprint{5099} & \numprint{834} & \numprint{48607} & \numprint{1967} & \numprint{7206} & \numprint{663} \\
10 Quantiles & \numprint{43151} & \numprint{964} & \numprint{5627} & \numprint{794} & \numprint{49120} & \numprint{2137} & \numprint{7682} & \numprint{673} \\
5 Quantiles & \numprint{43541} & \numprint{942} & \numprint{6131} & \numprint{1118} & \numprint{48846} & \numprint{2206} & \numprint{8233} & \numprint{836} \\
Top 15 & \numprint{43118} & \numprint{784} & \numprint{6458} & \numprint{793} & \numprint{48735} & \numprint{1923} & \numprint{28401} & \numprint{2365} \\
Top 30 & \numprint{43541} & \numprint{942} & \numprint{5068} & \numprint{472} & \numprint{48846} & \numprint{2206} & \numprint{28428} & \numprint{2385} \\
Top 60 & \numprint{44725} & \numprint{467} & \numprint{4634} & \numprint{305} & \numprint{48388} & \numprint{1521} & \numprint{28452} & \numprint{2493} \\

         \bottomrule
    \end{tabular}
\end{table}

Comparing the median GD\textsuperscript{+} and IGD\textsuperscript{+} of each obfuscation configuration between when two objectives are obfuscated (Table~\ref{tab:moap3_summary}) and when one objective is obfuscated (Table~\ref{tab:moap2_summary}) indicates that, except for \emph{order} obfuscation, median GD\textsuperscript{+} is lower when one objective is obfuscated, while, except for $A_{90}$, median IGD\textsuperscript{+} is lower when two objectives are obfuscated.
To this end, after 500 generations, obfuscating one objective tends to result in solution sets closer to the Pareto front whereas obfuscating two objectives tends to result in better coverage of the Pareto front.

\begin{figure}
    \centering
    \includegraphics[width=\linewidth]{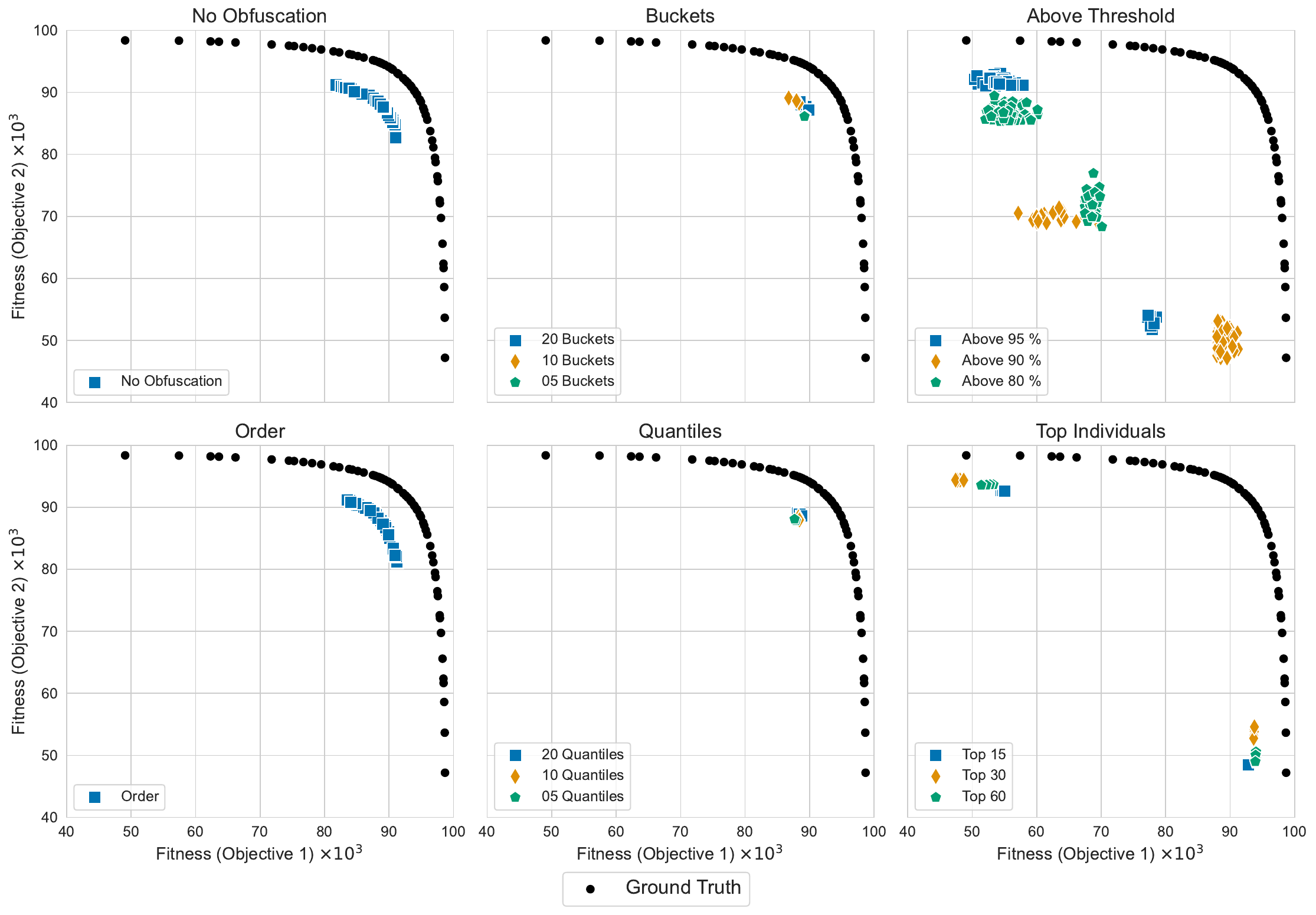}
    \caption{Set of estimated non-dominated solution sets of Generation 500 from the run with the median IGD\textsuperscript{+} for each obfuscation configuration on Instance \emph{ap2K100x100} with both objectives obfuscated.}
    \label{fig:moap3_ap2K100x100_results}
\end{figure}

Figure~\ref{fig:moap3_ap2K100x100_results} shows the set of estimated non-dominated solutions of Generation~500 from the run with the median IGD\textsuperscript{+} for each obfuscation configuration.
In contrast to the solution sets obtained when only Objective~1 is obfuscated, the solution sets are not skewed towards a single objective when both objectives are obfuscated.
The solution sets obtained by runs with \emph{fitness buckets} and \emph{order quantiles} obfuscation are in the center of the two objectives.
The solution sets obtained by the runs with \emph{above threshold} and \emph{top individuals} obfuscation tend to approach the extremes of both objectives.
These differences are because \emph{fitness buckets} and \emph{order quantiles} obfuscation reveal more information about the fitness of individuals in a population than \emph{above threshold} and \emph{top individuals} obfuscation.
Suppose the three candidate solutions $x$, $y$ and $z$ with a fitness $f(x)=(100,50)$, $f(y)=(50,100)$, and $f(z)=(60,60)$, then $z$ is likely to be below a threshold or not a top individual for both objectives, despite actually being non dominated.
With \emph{fitness buckets} and \emph{order quantiles} obfuscation, the buckets and quantile indices better reflect the actual fitness of $z$.

Table~\ref{tab:moap3_ap2K100x100_comparison_gd_plus} reports the median and IQR of the GD\textsuperscript{+} of the comparison generation, together with the results from comparing the GD\textsuperscript{+} of runs across obfuscation configurations.
The number of wins for each configuration is as follows: without obfuscation (13 wins), $O$ (11 wins), $T_{60}$ (11 wins), $T_{30}$ (10 wins), $A_{95}$ (8 wins), $Q_{10}$ (7 wins), $Q_{20}$ (6 wins), $Q_5$ (5 wins), $T_{15}$ (5 wins), $A_{90}$ (4 wins), $B_{20}$ (2 wins), $B_{10}$ (2 wins), $B_5$ (0 wins), and $A_{80}$ (0 wins).
The runs with $T_{60}$ and $T_{30}$ perform among the best, whereas those with \emph{fitness buckets} obfuscation perform among the worst.

\begin{table}
    \centering
    \caption{Median and interquartile range (IQR) of GD\textsuperscript{+} of the comparison generation across 31 runs for each obfuscation configuration on Instance \emph{ap2K100x100} with two objectives obfuscated. Matched-pairs rank-biserial correlations, rounded to one decimal place, are reported for comparing the performance of the configuration in each row with that in each column. Grey cells indicate statistically significant differences in performance. A dash (-) denotes no obfuscation.}
    \label{tab:moap3_ap2K100x100_comparison_gd_plus}
    \footnotesize
    \begin{tabular}{lrrrrrrrrrrrrrrr}
    \toprule
& \multirow{2}{*}{Median} & \multirow{2}{*}{IQR} & \multicolumn{13}{c}{Matched-Pairs Rank-Biserial Correlation} \\
\cmidrule(lr){4-16}
&        &     & - & $B_{20}$ & $B_{10}$ & $B_{5}$ & $A_{95}$ & $A_{90}$ & $A_{80}$ & $O$ & $Q_{20}$ & $Q_{10}$ & $Q_{5}$ & $T_{15}$ & $T_{30}$ \\
\midrule
- & \numprint{5597} & \numprint{546} \\
$B_{20}$ & \numprint{16823} & \numprint{810} & \cellcolor[HTML]{D3D3D3} \numprint{1.0} \\
$B_{10}$ & \numprint{16822} & \numprint{1214} & \cellcolor[HTML]{D3D3D3} \numprint{1.0} & -\numprint{0.0} \\
$B_{5}$ & \numprint{17893} & \numprint{1032} & \cellcolor[HTML]{D3D3D3} \numprint{1.0} & \cellcolor[HTML]{D3D3D3} \numprint{0.7} & \cellcolor[HTML]{D3D3D3} \numprint{0.8} \\
$A_{95}$ & \numprint{11393} & \numprint{1403} & \cellcolor[HTML]{D3D3D3} \numprint{1.0} & \cellcolor[HTML]{D3D3D3} -\numprint{1.0} & \cellcolor[HTML]{D3D3D3} -\numprint{1.0} & \cellcolor[HTML]{D3D3D3} -\numprint{1.0} \\
$A_{90}$ & \numprint{13950} & \numprint{1316} & \cellcolor[HTML]{D3D3D3} \numprint{1.0} & \cellcolor[HTML]{D3D3D3} -\numprint{1.0} & \cellcolor[HTML]{D3D3D3} -\numprint{1.0} & \cellcolor[HTML]{D3D3D3} -\numprint{1.0} & \cellcolor[HTML]{D3D3D3} \numprint{1.0} \\
$A_{80}$ & \numprint{17884} & \numprint{3237} & \cellcolor[HTML]{D3D3D3} \numprint{1.0} & \cellcolor[HTML]{D3D3D3} \numprint{0.5} & \cellcolor[HTML]{D3D3D3} \numprint{0.5} & \numprint{0.1} & \cellcolor[HTML]{D3D3D3} \numprint{1.0} & \cellcolor[HTML]{D3D3D3} \numprint{1.0} \\
$O$ & \numprint{9447} & \numprint{835} & \cellcolor[HTML]{D3D3D3} \numprint{1.0} & \cellcolor[HTML]{D3D3D3} -\numprint{1.0} & \cellcolor[HTML]{D3D3D3} -\numprint{1.0} & \cellcolor[HTML]{D3D3D3} -\numprint{1.0} & \cellcolor[HTML]{D3D3D3} -\numprint{1.0} & \cellcolor[HTML]{D3D3D3} -\numprint{1.0} & \cellcolor[HTML]{D3D3D3} -\numprint{1.0} \\
$Q_{20}$ & \numprint{11984} & \numprint{1022} & \cellcolor[HTML]{D3D3D3} \numprint{1.0} & \cellcolor[HTML]{D3D3D3} -\numprint{1.0} & \cellcolor[HTML]{D3D3D3} -\numprint{1.0} & \cellcolor[HTML]{D3D3D3} -\numprint{1.0} & \cellcolor[HTML]{D3D3D3} \numprint{0.6} & \cellcolor[HTML]{D3D3D3} -\numprint{1.0} & \cellcolor[HTML]{D3D3D3} -\numprint{1.0} & \cellcolor[HTML]{D3D3D3} \numprint{1.0} \\
$Q_{10}$ & \numprint{11884} & \numprint{1247} & \cellcolor[HTML]{D3D3D3} \numprint{1.0} & \cellcolor[HTML]{D3D3D3} -\numprint{1.0} & \cellcolor[HTML]{D3D3D3} -\numprint{1.0} & \cellcolor[HTML]{D3D3D3} -\numprint{1.0} & \numprint{0.4} & \cellcolor[HTML]{D3D3D3} -\numprint{0.9} & \cellcolor[HTML]{D3D3D3} -\numprint{1.0} & \cellcolor[HTML]{D3D3D3} \numprint{1.0} & -\numprint{0.1} \\
$Q_{5}$ & \numprint{12592} & \numprint{1095} & \cellcolor[HTML]{D3D3D3} \numprint{1.0} & \cellcolor[HTML]{D3D3D3} -\numprint{1.0} & \cellcolor[HTML]{D3D3D3} -\numprint{1.0} & \cellcolor[HTML]{D3D3D3} -\numprint{1.0} & \cellcolor[HTML]{D3D3D3} \numprint{0.8} & \cellcolor[HTML]{D3D3D3} -\numprint{0.8} & \cellcolor[HTML]{D3D3D3} -\numprint{1.0} & \cellcolor[HTML]{D3D3D3} \numprint{1.0} & \cellcolor[HTML]{D3D3D3} \numprint{0.5} & \cellcolor[HTML]{D3D3D3} \numprint{0.5} \\
$T_{15}$ & \numprint{12335} & \numprint{1028} & \cellcolor[HTML]{D3D3D3} \numprint{1.0} & \cellcolor[HTML]{D3D3D3} -\numprint{1.0} & \cellcolor[HTML]{D3D3D3} -\numprint{1.0} & \cellcolor[HTML]{D3D3D3} -\numprint{1.0} & \cellcolor[HTML]{D3D3D3} \numprint{0.7} & \cellcolor[HTML]{D3D3D3} -\numprint{0.9} & \cellcolor[HTML]{D3D3D3} -\numprint{1.0} & \cellcolor[HTML]{D3D3D3} \numprint{1.0} & \numprint{0.4} & \cellcolor[HTML]{D3D3D3} \numprint{0.5} & -\numprint{0.1} \\
$T_{30}$ & \numprint{10406} & \numprint{726} & \cellcolor[HTML]{D3D3D3} \numprint{1.0} & \cellcolor[HTML]{D3D3D3} -\numprint{1.0} & \cellcolor[HTML]{D3D3D3} -\numprint{1.0} & \cellcolor[HTML]{D3D3D3} -\numprint{1.0} & \cellcolor[HTML]{D3D3D3} -\numprint{0.8} & \cellcolor[HTML]{D3D3D3} -\numprint{1.0} & \cellcolor[HTML]{D3D3D3} -\numprint{1.0} & \cellcolor[HTML]{D3D3D3} \numprint{0.8} & \cellcolor[HTML]{D3D3D3} -\numprint{1.0} & \cellcolor[HTML]{D3D3D3} -\numprint{1.0} & \cellcolor[HTML]{D3D3D3} -\numprint{1.0} & \cellcolor[HTML]{D3D3D3} -\numprint{1.0} \\
$T_{60}$ & \numprint{9582} & \numprint{708} & \cellcolor[HTML]{D3D3D3} \numprint{1.0} & \cellcolor[HTML]{D3D3D3} -\numprint{1.0} & \cellcolor[HTML]{D3D3D3} -\numprint{1.0} & \cellcolor[HTML]{D3D3D3} -\numprint{1.0} & \cellcolor[HTML]{D3D3D3} -\numprint{1.0} & \cellcolor[HTML]{D3D3D3} -\numprint{1.0} & \cellcolor[HTML]{D3D3D3} -\numprint{1.0} & \numprint{0.1} & \cellcolor[HTML]{D3D3D3} -\numprint{1.0} & \cellcolor[HTML]{D3D3D3} -\numprint{1.0} & \cellcolor[HTML]{D3D3D3} -\numprint{1.0} & \cellcolor[HTML]{D3D3D3} -\numprint{1.0} & \cellcolor[HTML]{D3D3D3} -\numprint{0.9} \\

    \bottomrule
    \end{tabular}
\end{table}

Table~\ref{tab:moap3_ap2K100x100_comparison_igd_plus} reports the median and IQR of the IGD\textsuperscript{+} of the comparison generation, together with the results from comparing the IGD\textsuperscript{+} of runs across obfuscation configurations.
The number of wins for each configuration is as follows: without obfuscation (13 wins), $O$ (12 wins), $Q_{20}$ (11 wins), $Q_{10}$ (10 wins), $Q_5$ (9 wins), $B_{20}$ (7 wins), $B_{10}$ (7 wins), $B_5$ (6 wins), $A_{80}$ (4 wins), $T_{60}$ (3 wins), $A_{95}$ (1 win), $T_{30}$ (1 win), $A_{90}$ (0 wins), and $T_{15}$ (0 wins).
Among the configurable obfuscation methods, the runs with \emph{order quantiles} obfuscation perform best, followed by those with \emph{fitness buckets}.
The runs with \emph{above threshold} and \emph{top individuals} obfuscation perform worst.

\begin{table}
    \centering
    \caption{Median and interquartile range (IQR) of IGD\textsuperscript{+} of the comparison generation across 31 runs for each obfuscation configuration on Instance \emph{ap2K100x100} with two objectives obfuscated. Matched-pairs rank-biserial correlations, rounded to one decimal place, are reported for comparing the performance of the configuration in each row with that in each column. Grey cells indicate statistically significant differences in performance. A dash (-) denotes no obfuscation.}
    \label{tab:moap3_ap2K100x100_comparison_igd_plus}
    \footnotesize
    \begin{tabular}{lrrrrrrrrrrrrrrr}
    \toprule
& \multirow{2}{*}{Median} & \multirow{2}{*}{IQR} & \multicolumn{13}{c}{Matched-Pairs Rank-Biserial Correlation} \\
\cmidrule(lr){4-16}
&        &     & - & $B_{20}$ & $B_{10}$ & $B_{5}$ & $A_{95}$ & $A_{90}$ & $A_{80}$ & $O$ & $Q_{20}$ & $Q_{10}$ & $Q_{5}$ & $T_{15}$ & $T_{30}$ \\
\midrule
- & \numprint{5951} & \numprint{566} \\
$B_{20}$ & \numprint{17606} & \numprint{1145} & \cellcolor[HTML]{D3D3D3} \numprint{1.0} \\
$B_{10}$ & \numprint{17253} & \numprint{1612} & \cellcolor[HTML]{D3D3D3} \numprint{1.0} & -\numprint{0.1} \\
$B_{5}$ & \numprint{18728} & \numprint{1026} & \cellcolor[HTML]{D3D3D3} \numprint{1.0} & \cellcolor[HTML]{D3D3D3} \numprint{0.8} & \cellcolor[HTML]{D3D3D3} \numprint{0.8} \\
$A_{95}$ & \numprint{27865} & \numprint{1845} & \cellcolor[HTML]{D3D3D3} \numprint{1.0} & \cellcolor[HTML]{D3D3D3} \numprint{1.0} & \cellcolor[HTML]{D3D3D3} \numprint{1.0} & \cellcolor[HTML]{D3D3D3} \numprint{1.0} \\
$A_{90}$ & \numprint{28312} & \numprint{2827} & \cellcolor[HTML]{D3D3D3} \numprint{1.0} & \cellcolor[HTML]{D3D3D3} \numprint{1.0} & \cellcolor[HTML]{D3D3D3} \numprint{1.0} & \cellcolor[HTML]{D3D3D3} \numprint{1.0} & \numprint{0.1} \\
$A_{80}$ & \numprint{26439} & \numprint{2680} & \cellcolor[HTML]{D3D3D3} \numprint{1.0} & \cellcolor[HTML]{D3D3D3} \numprint{1.0} & \cellcolor[HTML]{D3D3D3} \numprint{1.0} & \cellcolor[HTML]{D3D3D3} \numprint{1.0} & \cellcolor[HTML]{D3D3D3} -\numprint{0.6} & \cellcolor[HTML]{D3D3D3} -\numprint{0.7} \\
$O$ & \numprint{9762} & \numprint{706} & \cellcolor[HTML]{D3D3D3} \numprint{1.0} & \cellcolor[HTML]{D3D3D3} -\numprint{1.0} & \cellcolor[HTML]{D3D3D3} -\numprint{1.0} & \cellcolor[HTML]{D3D3D3} -\numprint{1.0} & \cellcolor[HTML]{D3D3D3} -\numprint{1.0} & \cellcolor[HTML]{D3D3D3} -\numprint{1.0} & \cellcolor[HTML]{D3D3D3} -\numprint{1.0} \\
$Q_{20}$ & \numprint{12605} & \numprint{738} & \cellcolor[HTML]{D3D3D3} \numprint{1.0} & \cellcolor[HTML]{D3D3D3} -\numprint{1.0} & \cellcolor[HTML]{D3D3D3} -\numprint{1.0} & \cellcolor[HTML]{D3D3D3} -\numprint{1.0} & \cellcolor[HTML]{D3D3D3} -\numprint{1.0} & \cellcolor[HTML]{D3D3D3} -\numprint{1.0} & \cellcolor[HTML]{D3D3D3} -\numprint{1.0} & \cellcolor[HTML]{D3D3D3} \numprint{1.0} \\
$Q_{10}$ & \numprint{13218} & \numprint{1089} & \cellcolor[HTML]{D3D3D3} \numprint{1.0} & \cellcolor[HTML]{D3D3D3} -\numprint{1.0} & \cellcolor[HTML]{D3D3D3} -\numprint{1.0} & \cellcolor[HTML]{D3D3D3} -\numprint{1.0} & \cellcolor[HTML]{D3D3D3} -\numprint{1.0} & \cellcolor[HTML]{D3D3D3} -\numprint{1.0} & \cellcolor[HTML]{D3D3D3} -\numprint{1.0} & \cellcolor[HTML]{D3D3D3} \numprint{1.0} & \cellcolor[HTML]{D3D3D3} \numprint{0.4} \\
$Q_{5}$ & \numprint{13650} & \numprint{1157} & \cellcolor[HTML]{D3D3D3} \numprint{1.0} & \cellcolor[HTML]{D3D3D3} -\numprint{1.0} & \cellcolor[HTML]{D3D3D3} -\numprint{1.0} & \cellcolor[HTML]{D3D3D3} -\numprint{1.0} & \cellcolor[HTML]{D3D3D3} -\numprint{1.0} & \cellcolor[HTML]{D3D3D3} -\numprint{1.0} & \cellcolor[HTML]{D3D3D3} -\numprint{1.0} & \cellcolor[HTML]{D3D3D3} \numprint{1.0} & \cellcolor[HTML]{D3D3D3} \numprint{0.8} & \cellcolor[HTML]{D3D3D3} \numprint{0.5} \\
$T_{15}$ & \numprint{28827} & \numprint{1465} & \cellcolor[HTML]{D3D3D3} \numprint{1.0} & \cellcolor[HTML]{D3D3D3} \numprint{1.0} & \cellcolor[HTML]{D3D3D3} \numprint{1.0} & \cellcolor[HTML]{D3D3D3} \numprint{1.0} & \cellcolor[HTML]{D3D3D3} \numprint{0.4} & \numprint{0.2} & \cellcolor[HTML]{D3D3D3} \numprint{0.8} & \cellcolor[HTML]{D3D3D3} \numprint{1.0} & \cellcolor[HTML]{D3D3D3} \numprint{1.0} & \cellcolor[HTML]{D3D3D3} \numprint{1.0} & \cellcolor[HTML]{D3D3D3} \numprint{1.0} \\
$T_{30}$ & \numprint{27694} & \numprint{1876} & \cellcolor[HTML]{D3D3D3} \numprint{1.0} & \cellcolor[HTML]{D3D3D3} \numprint{1.0} & \cellcolor[HTML]{D3D3D3} \numprint{1.0} & \cellcolor[HTML]{D3D3D3} \numprint{1.0} & -\numprint{0.1} & -\numprint{0.2} & \cellcolor[HTML]{D3D3D3} \numprint{0.5} & \cellcolor[HTML]{D3D3D3} \numprint{1.0} & \cellcolor[HTML]{D3D3D3} \numprint{1.0} & \cellcolor[HTML]{D3D3D3} \numprint{1.0} & \cellcolor[HTML]{D3D3D3} \numprint{1.0} & \cellcolor[HTML]{D3D3D3} -\numprint{0.5} \\
$T_{60}$ & \numprint{26886} & \numprint{2917} & \cellcolor[HTML]{D3D3D3} \numprint{1.0} & \cellcolor[HTML]{D3D3D3} \numprint{1.0} & \cellcolor[HTML]{D3D3D3} \numprint{1.0} & \cellcolor[HTML]{D3D3D3} \numprint{1.0} & \cellcolor[HTML]{D3D3D3} -\numprint{0.4} & \cellcolor[HTML]{D3D3D3} -\numprint{0.6} & \numprint{0.2} & \cellcolor[HTML]{D3D3D3} \numprint{1.0} & \cellcolor[HTML]{D3D3D3} \numprint{1.0} & \cellcolor[HTML]{D3D3D3} \numprint{1.0} & \cellcolor[HTML]{D3D3D3} \numprint{1.0} & \cellcolor[HTML]{D3D3D3} -\numprint{0.7} & -\numprint{0.4} \\

    \bottomrule
    \end{tabular}
\end{table}

Compared to solution sets obtained when obfuscating one objective, when both objectives are obfuscated, solution sets tend to be slightly more distant from the Pareto front while providing better coverage.
Although reducing information leakage, the results show that obfuscation indeed negatively impacts GD\textsuperscript{+} and IGD\textsuperscript{+} when imposing the time budged on evaluation.
As with obfuscation of only one objective, obfuscation configurations that perform well on GD\textsuperscript{+} may perform poorly on IGD\textsuperscript{+}, and vice versa.
When summing the comparison wins of GD\textsuperscript{+} and IGD\textsuperscript{+}, runs with $Q_{20}$ and $Q_{10}$ perform best among the configurable obfuscation methods within the time budget.

\section{Conclusion}\label{sec:conclusion}

In this paper, we presented an approach for privacy-preserving distributed optimization in time-critical settings.
Our approach uses evolutionary algorithms for solution search and MPC only for the evaluation of the solutions to reduce the runtime overhead of privacy-preserving computations.
Obfuscation methods additionally reduce the information revealed to an honest-but-curious platform provider, which introduces a potential trade-off between protection of private information and solution quality.
We experimentally evaluated the impact of obfuscation on the performance of evolutionary algorithms.
The experimental results demonstrate that a genetic algorithm is capable of finding near-optimal solutions for the AP and the TSP.
NSGA-II closely approximates the Pareto front, but with poor coverage when only one objective is obfuscated on the MOAP.
Obfuscating both objectives slightly increases the distance to the Pareto front but improves coverage.
For each problem class, we additionally compared the performance of evolutionary algorithms across obfuscation configurations based on a fixed time budget for evaluation to ensure comparability.

There are several aspects to be further investigated in future work.
First, we will investigate the impact of obfuscation on performance without revealing the maximum fitness of populations to further protect optimization preferences from an honest-but-curious platform provider.
Second, we will further explore multi-objective optimization, where a single Privacy Engine instance evaluates multiple objectives and returns obfuscated evaluation results, e.g., based on Pareto rank and crowding distance for NSGA-II.
This may, however, come with a strong runtime penalty due to additional privacy-preserving computations.
Finally, we will consider long-term equity among participants in distributed optimization problems.
Participants may accept being disadvantaged in individual optimization instances to improve overall efficiency, but no participant should have an advantage or disadvantage compared to others over multiple instances.

\begin{acks}
This work was conducted as part of the HARMONIC project. 
This project has received funding from the SESAR Joint Undertaking under grant agreement No 101114675 under the European Union's Horizon Europe research and innovation program. 
UK participant NATS in Horizon Europe Project HARMONIC receives funding from UK Research and Innovation (UKRI) under the UK government’s Horizon Europe funding guarantee [grant number 10091990].
This work has received funding from the Swiss State Secretariat for Education, Research and Innovation (SERI).
The views expressed in this paper are those of the authors.

\bigskip

\begin{center}
	\raisebox{25pt}{\includegraphics[width=.24\textwidth]{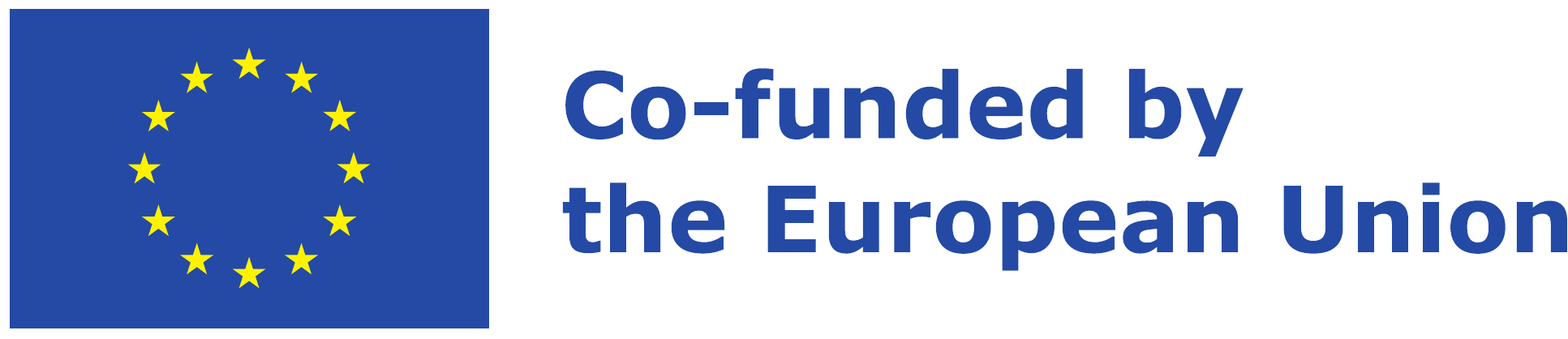}}
	\hfill
	\raisebox{25pt}{\includegraphics[width=.12\textwidth]{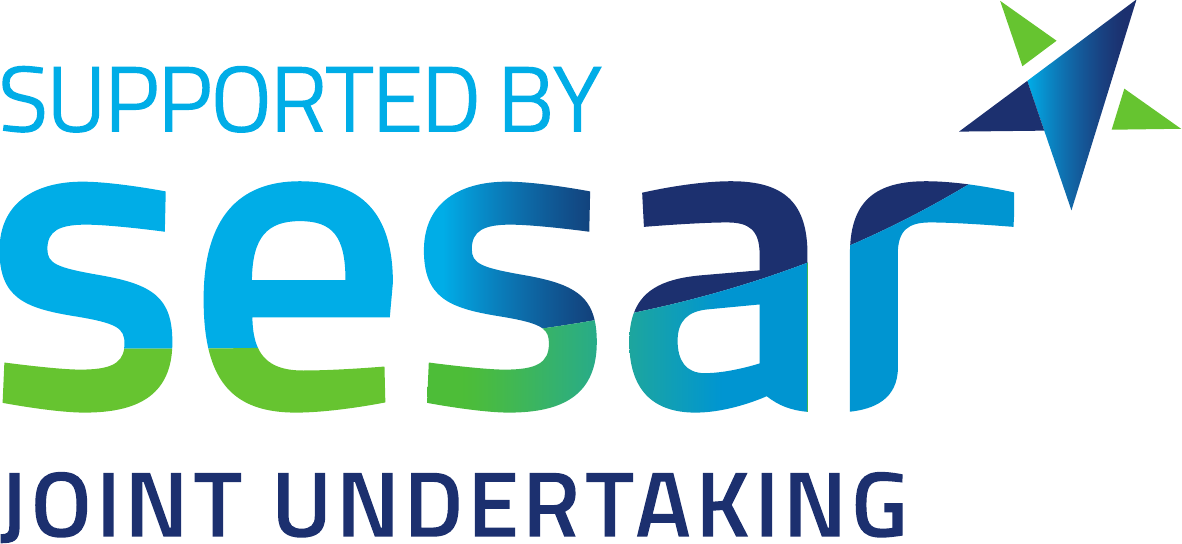}}
	\hfill
	\raisebox{0pt}{\includegraphics[width=.48\textwidth]{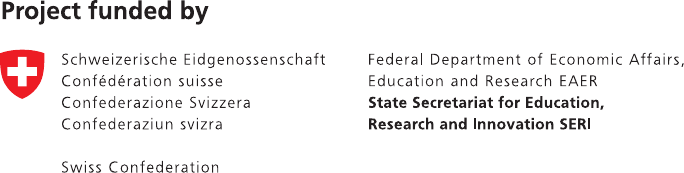}}
\end{center}

\end{acks}

\bibliographystyle{ACM-Reference-Format}
\bibliography{references}

\end{document}